\documentclass{ecai} 

\usepackage{amsmath, amssymb, mathtools, amsthm}
\usepackage{latexsym}

\usepackage{booktabs}
\usepackage{tabularx}
\usepackage{multirow}
\usepackage{enumitem}

\newcolumntype{R}{>{\raggedleft\arraybackslash}X}
\newcolumntype{L}{>{\raggedright\arraybackslash}X}
\newcolumntype{C}{>{\centering\arraybackslash}X}

\usepackage{graphicx} 
\usepackage{color}

\usepackage[capitalize,noabbrev]{cleveref}

\usepackage[nolist,nohyperlinks]{acronym}  

\usepackage{pifont}  

\newcommand{\Pos}{\mathcal{P}}
\newcommand{\B}{\mathcal{B}}

\newcommand{\sd}[1]{\text{\tiny{}\textpm#1}}

\DeclareMathOperator{\NTXent}{\textup{NT-Xent}}
\DeclareMathOperator{\FIRM}{\textup{FIRM}}
\newcommand{\scon}{s_{\textup{con}}}
\newcommand{\sshift}{s_{\textup{shift}}}
\newcommand{\sens}{s_{\textup{ens}}}

\makeatletter
\newcommand*\bigcdot{\mathpalette\bigcdot@{.5}}
\newcommand*\bigcdot@[2]{\mathbin{\vcenter{\hbox{\scalebox{#2}{$\m@th#1\bullet$}}}}}
\makeatother

\newcommand{\BibTeX}{B\kern-.05em{\sc i\kern-.025em b}\kern-.08em\TeX}

\usepackage{xcolor}

\definecolor{myred}{HTML}{DC3912} 
\definecolor{myblue}{HTML}{3366CC} 
\definecolor{mygreen}{HTML}{109618} 

\newcommand{\showdiff}[1]{%
    \ifdim#1pt>0pt\relax
        \color{mygreen}+#1\%  
    \else
        \color{myred}#1\%     
    \fi
}

\begin{document}

\begin{acronym} 
    \acro{AUROC}{Area Under the ROC Curve}
    \acro{AULC}{Area Under the Learning Curve}
    \acro{FIRM}{Focused In-distribution Representation Modeling}
    \acro{ID}{In-Distribution}
    \acro{KDE}{Kernel Density Estimation}
    \acro{OE}{Outlier Exposure}
    \acro{OOD}{Out-of-distribution}
\end{acronym} 

\begin{frontmatter}

\title{Contrastive Representation Modeling \\ for Anomaly Detection}

\author[A]{\fnms{Willian T.}~\snm{Lunardi}\thanks{Corresponding author. Email: willian.lunardi@tii.ae. \\ {Accepted for publication at the 28th European Conference on Artificial Intelligence (ECAI 2025).}} }
\author[A]{\fnms{Abdulrahman}~\snm{Banabila}}
\author[A]{\fnms{Dania}~\snm{Herzalla}}
\author[A,B]{\fnms{Martin}~\snm{Andreoni}}

\address[A]{Technology Innovation Institute, Abu Dhabi, UAE}
\address[B]{Computer Science Department, Khalifa University of Science and Technology, Abu Dhabi, UAE}

\begin{abstract}
    Distance-based anomaly detection methods rely on compact in-distribution (ID) embeddings that are well separated from anomalies. However, conventional contrastive learning strategies often struggle to achieve this balance, either promoting excessive variance among inliers or failing to preserve the diversity of outliers. We begin by analyzing the challenges of representation learning for anomaly detection and identify three essential properties for the pretext task: (1) compact clustering of inliers, (2) strong separation between inliers and anomalies, and (3) preservation of diversity among synthetic outliers. Building on this, we propose a structured contrastive objective that redefines positive and negative relationships during training, promoting these properties without requiring explicit anomaly labels. We extend this framework with a patch-based learning and evaluation strategy specifically designed to improve the detection of localized anomalies in industrial settings. Our approach demonstrates significantly faster convergence and improved performance compared to standard contrastive methods. It matches or surpasses anomaly detection methods on both semantic and industrial benchmarks, including methods that rely on discriminative training or explicit anomaly labels. 
\end{abstract}

\end{frontmatter}

\section{Introduction}
Anomaly detection is crucial across numerous domains, including cybersecurity, healthcare, and autonomous systems, where identifying rare or unusual patterns is essential for ensuring robustness, data integrity, and failure prevention. This task is closely related to fields such as \ac{OOD} detection, novelty detection, and one-class classification (i.e., semantic anomaly detection), ll of which operate within a specific \ac{ID} setting to identify data points that do not conform to expected patterns under the open-world assumption~\citep{yang2024generalized}. While OOD detection typically involves distinguishing between predefined classes with labeled datasets, anomaly detection focuses on identifying deviations from a singular normal class, where minimizing intra-class variance and ensuring tightly clustered inlier representations are particularly crucial for distance-based methods to effectively separate anomalies from normal samples~\citep{ming2023cider}.

Representation learning plays a critical role in anomaly detection, with many approaches relying on distance-based scores, such as cosine similarity, computed in the learned representation~\citep{ruff2018deep,ruff2019deep,csi,sehwag2021ssd,sun2022out,reiss2023mean}. These methods assume that OOD samples are situated far from the compact clusters formed by inlier data, allowing for the classification of anomalies based on their relative positioning. However, their effectiveness is contingent on the quality of the learned embeddings~\citep{ming2023cider}, as compact and well-separated ID representations are essential for clear anomaly separation.

A key challenge in anomaly detection is structuring learning objectives to balance inlier compactness and outlier separation. Prior work has introduced synthetic outliers to improve anomaly separation, generated through methods such as geometric transformations~\citep{golan2018deep,hendrycks2019using}, local perturbations~\citep{li2021cutpaste,schluter2022natural}, external datasets~\citep{outlier_exposure}, and generative outlier synthesis~\cite{du2024dream}. However, existing contrastive formulations do not fully leverage these methods, as they often fail to incorporate both inlier structure and outlier diversity into the learning objective. The single-positive formulation does not effectively capture the structure of inlier data, leading to increased variance within the normal class. Naive multi-positive extensions fail to capture the diversity among outliers, weakening contrastive signals and hindering effective learning.

We begin by diagnosing a fundamental weakness in existing contrastive formulations for anomaly detection, demonstrating that an effective contrastive objective must explicitly enforce three key properties: (1) compact inlier clustering, (2) strong separation between inliers and synthetic outliers, and (3) preserve diversity among synthetic outliers to maintain meaningful contrastive signals. Building on this insight, we reconceptualize how contrastive learning should be applied to anomaly detection by formulating a structured approach that directly integrates these properties. To this end, we propose \ac{FIRM}, a representation learning framework that enforces these essential properties to ensure more effective anomaly separation. \ac{FIRM} explicitly promotes compact clustering of \ac{ID} representations, reducing intra-class variance and structuring inlier embeddings for improved detection. Simultaneously, it enhances the separation between inliers and synthetic outliers while preserving the diversity among synthetic outliers, preventing representation collapse and ensuring that anomalous variations remain distinct. In summary, our contributions include:
\begin{itemize}
    \item We analyze the limitations of contrastive learning for anomaly detection, particularly the issue of class collision, and formalize key properties for effective anomaly separation.
    \item We introduce \ac{FIRM}, a contrastive learning objective that encourages inlier compactness, enhances inlier-outlier separation, and preserves synthetic outlier diversity.
    \item We extend FIRM to a patch-based contrastive learning framework for industrial anomaly detection, leveraging foreground-aware sampling and realistic synthetic anomaly injection to enable fine-grained localization without pixel-level supervision.
    \item We validate \ac{FIRM} on both semantic and industrial anomaly detection tasks, showing substantial performance and convergence improvements. Notably, in industrial settings, our unsupervised patch-based method surpasses several pixel-level supervised approaches, despite requiring no localization labels.
\end{itemize}

The remainder of this paper is organized as follows: Section~\ref{sec:related_work} reviews relevant literature on contrastive learning and anomaly detection. Section~\ref{sec:learning_robust} introduces the formal problem and examines challenges in existing contrastive methods. In Section~\ref{sec:multi_positive}, we present our proposed contrastive objective, followed by Section~\ref{sec:patch_learning}, which describes our patch-based learning strategy for industrial anomalies. Finally, Section~\ref{sec:experiments} reports experimental results across multiple benchmarks.

\section{Related Work}\label{sec:related_work}

\paragraph{Self-supervised learning.} 

Self-supervised learning~\citep{rotpred,hendrycks2019using}, particularly contrastive learning~\citep{oord2018representation} via instance discrimination~\citep{wu2018unsupervised}, has been widely adopted for learning representations without labeled data~\citep{simclr,moco,mocov2,mocov3}. Contrastive objectives like InfoNCE~\citep{oord2018representation} and NT-Xent~\citep{simclr} align positives while separating negatives through self-labeling. SupCon~\citep{supcon,supcon2} extends this by leveraging labels to cluster positives more tightly and enhance class separation.

\paragraph{Representation learning for anomaly detection.}  
The limitations of classical anomaly detection approaches have driven significant interest in developing deep learning methods for anomaly detection and related tasks~\citep{yi2020psvdd,li2021cutpaste,zavrtanik2021draem,roth2022towards,zhang2023destseg,zhang2023unsupervised}. Most of these approaches operate in an unsupervised setting~\citep{ruff2018deep,golan2018deep}. However, incorporating additional data as negatives can significantly enhance performance~\citep{hendrycks2019using,outlier_exposure,du2022vos,du2024dream}. Recent works have explored the application of contrastive learning for anomaly detection~\citep{learn_and_eval}, particularly in \ac{OOD} detection~\citep{csi,sehwag2021ssd,sun2022out,ming2023cider,wang2023learning} and open-set domains~\citep{bucci2022distance}, with several studies proposing contrastive formulations for improved detection. \citet{reiss2023mean} adapts the contrastive loss but relies on pre-trained models. \citet{ming2023cider} emphasizes compactness and dispersion in labeled \ac{OOD} detection, while \citet{wang2023uniconha} incorporates hierarchical augmentations and re-weighting to refine inlier concentration and outlier dispersion. In contrast, we take a principled approach by systematically examining how the definition and expansion of the positive set influences representation structure and propose a formulation that aligns with these key properties for effective anomaly separation.

\begin{figure}[t]
    \centering
    \includegraphics[width=0.99\linewidth]{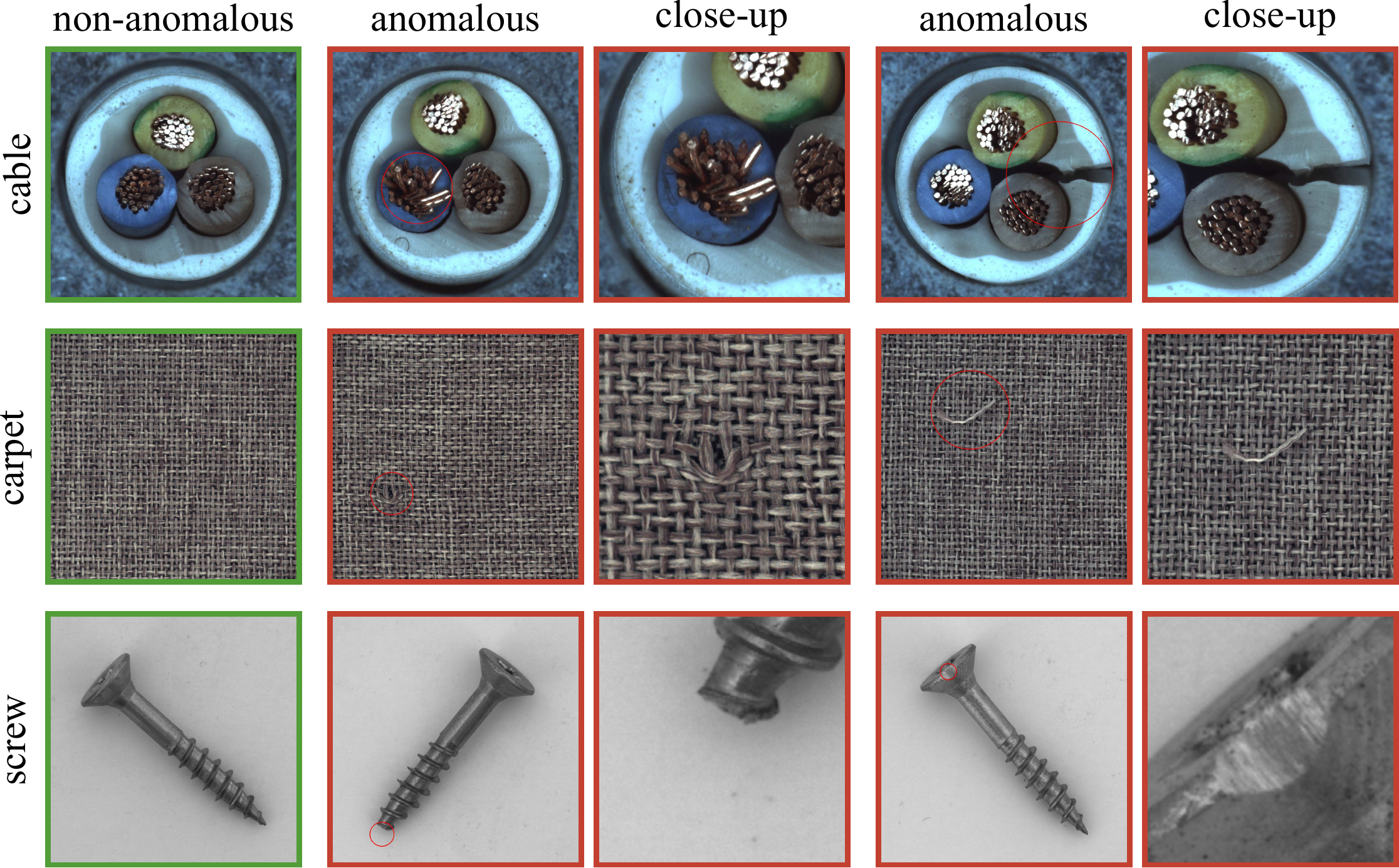} 
    \caption{Diversity of anomalies in real-world anomaly detection tasks. Anomalies can manifest in various forms, including scratches, deformations, and missing parts. This variability is reflected in the generation process (often stochastic) and underscores the challenge of treating synthetic outliers as a homogeneous group in representation learning.}
    \label{fig:mvtec_anomalies}
\end{figure}

\section{Normality Manifold Learning}\label{sec:learning_robust} 

\subsection{Problem Formulation}
Consider a dataset \(\mathcal{D}_{\textup{in}} = \{x_i\}_{i=1}^N\) consisting of \(N\) normal samples, where each \(x_i \in \mathcal{X}\) is drawn independently and identically distributed from the \ac{ID} distribution \(P_{\textup{in}}\). Here, \(P_{\textup{in}}\) represents the underlying distribution of normal data. For clarity, we define the label space as \(\mathcal{Y} \in \{1, -1\}\), where \(y_i = 1\) denotes \ac{ID} samples (normals), and \(y_i = -1\) represents anomalies (OOD samples). Since \(\mathcal{D}_{\textup{in}}\) contains only normal data, all samples within it satisfy \(y_i = 1\). The objective is to learn an encoder \(f_{\theta}: \mathcal{X} \to \mathbb{R}^d\), parameterized by \(\theta\), that maps \(\mathcal{D}_{\textup{in}}\) to a compact representation in \(\mathbb{R}^d\), ensuring that anomalies are projected to distinct and dispersed regions in the representation space.

Once trained, anomaly detection is framed as a binary classification task. The scoring function \(S(x)\) assigns a score to each sample \(x \in \mathcal{X}\) based on the encoder's output \(f_\theta(x)\). In this work, we consider non-parametric scoring functions, such as distance metrics computed between \(f_\theta(x)\) and reference embeddings derived from \ac{ID} samples. The scoring function \(S(x)\) is thresholded to classify a sample as anomalous. Specifically, a sample \(x\) is classified as an anomaly if \(S(x) > \tau,\) where \(\tau\) is a predefined detection threshold. The binary label \(y \in \mathcal{Y}\) is assigned based on the thresholding rule, with \(y = 1\) indicating that the sample belongs to the \ac{ID} distribution and \(y = -1\) indicating that it is anomalous (OOD).

For clarity, we use \textit{inliers} to refer to normal samples drawn from the \textit{in-distribution} \(P_{\textup{in}}\), and \textit{out-of-distribution} (OOD) to refer to any outlier sample outside \(P_{\textup{in}}\), including real anomalies and synthetic outliers.

\begin{figure*}[!t]
    \scriptsize
    \centering
    \setlength{\tabcolsep}{0pt}
    \begin{tabular}{cccc} \includegraphics[width=0.25\textwidth]{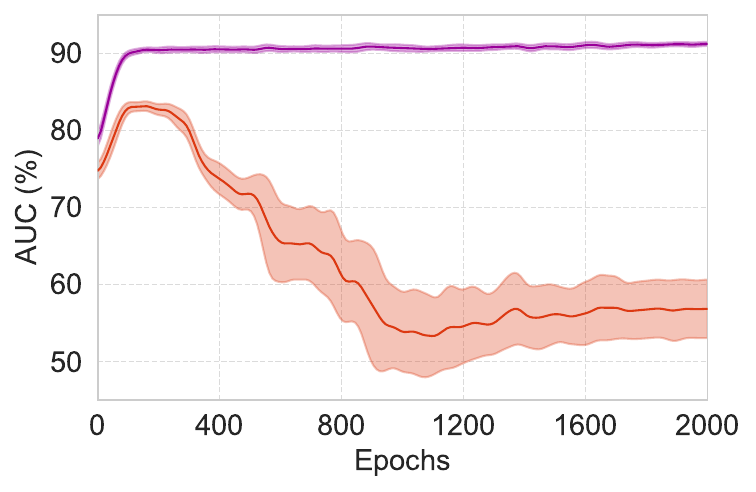} &  \includegraphics[width=0.25\textwidth]{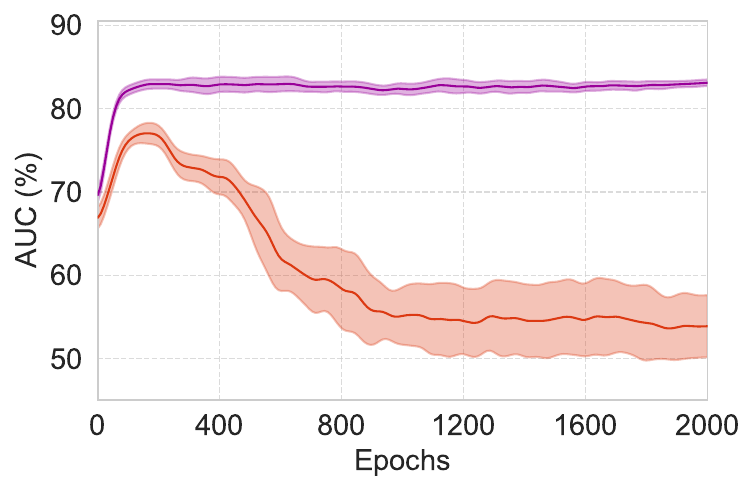} &  \includegraphics[width=0.25\textwidth]{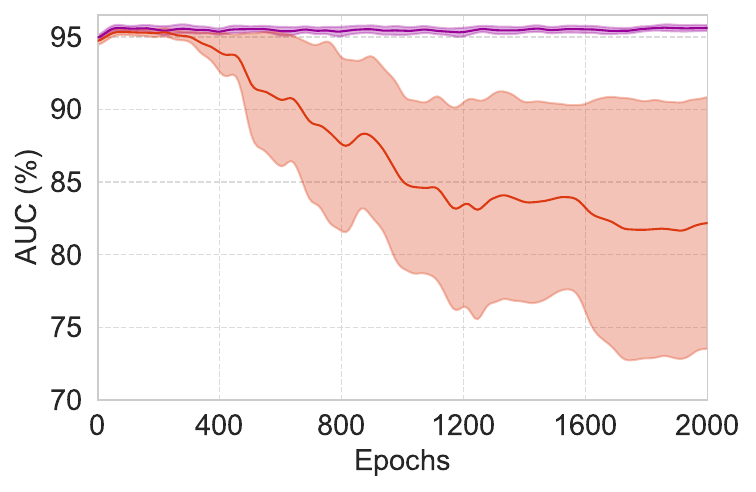} &  \includegraphics[width=0.25\textwidth]{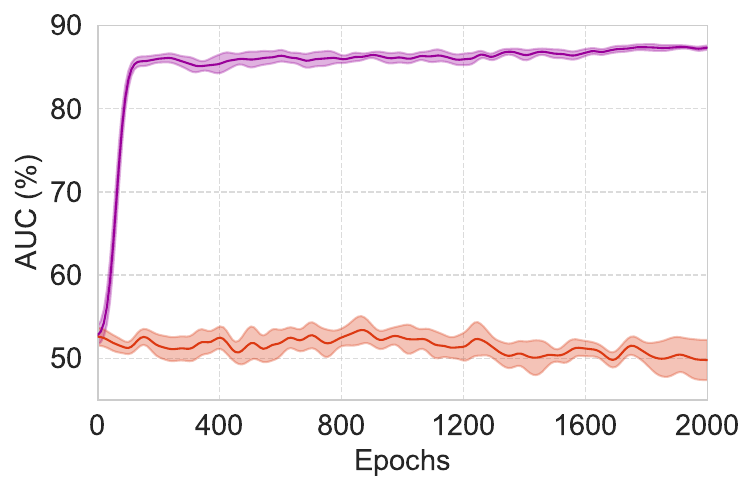}  \\
        (a) CIFAR-10. & (b) CIFAR-100 (superclass). & (c) Fashion-MNIST.  & (d) Cats-vs-Dogs.
    \end{tabular} 
    \caption{Effect of defining $\Pos(i)$ for synthetic outliers by including either all other synthetic outliers in the batch (red) or only those with the same transformation type (purple). Enforcing separation between synthetic outliers prevents representation collapse and improves anomaly detection.}\label{fig:supcon_comparison}
\end{figure*}

\subsection{Learning through Contrastive Objectives}
Self-supervised learning strategies, such as contrastive representation learning, optimize a loss function to bring representations of semantically similar samples closer together while pushing dissimilar ones farther apart.
For a given unlabeled sample $x_i \sim \mathcal{X}$, a stochastic data augmentation function $\alpha$ is applied to create two correlated \textit{instances}, denoted as a positive pair $(\tilde{x}_i, \tilde{x}_{i^+})$. In a minibatch of $n$ samples, the augmentation process leads to a multiview minibatch $\mathcal{B} = \{1, \ldots, 2n\}$. Within this multiview batch, each instance  $\tilde x_i$ and its counterpart $\tilde x_{i^+}$ serve as the \textit{anchor} and the \textit{positive} respectively. At the same time, all other samples are treated as \textit{negatives}.

Each instance within $\B$ is encoded via a neural network encoder $f_{\theta}$ into embeddings $f_{\theta}(\tilde{x}_i) \in \mathbb{R}^d$. The encoder's output is further transformed by a projection network $g_{\psi}$, parametrized by $\psi$, into a lower-dimensional space, resulting in vectors $z_i = g_{\psi}(f_{\theta}(\tilde{x}_i)) \in \mathbb{R}^{d_{\textup{head}}}$ where $d_{\textup{head}} < d$~\citep{rotpred}.
The core of the learning process is driven by a contrastive objective function~\citep{sohn2016improved, oord2018representation, simclr}, which encourages the maximization of the similarity between the representations of the anchor and the positive while minimizing similarity to negatives. Following~\citep{simclr}, the training objective takes the following form:
\begin{equation}\label{eq:ntxent}
    \mathcal{L}_{\NTXent}(\B) = -\sum_{i \in \mathcal{B}} \log \frac{\exp(z_i \bigcdot z_{i^+} / \tau)}{\sum_{a \in \mathcal{B} \setminus \{i\}} \exp(z_i \bigcdot z_a / \tau)},
\end{equation}
where $z_i = g_{\psi}(f_{\theta}(\tilde{x}_i)) / \|g_{\psi}(f_{\theta}(\tilde{x}_i))\|$ are the normalized outputs of the projection network $g_{\psi}$, symbol $\bigcdot$ denotes the dot product, and $\tau$ is a positive scalar known as the temperature parameter.  

\subsection{Class Collision in Contrastive Learning}\label{sec:class_collision}
Contrastive learning relies on negative samples to shape the representation space effectively. For a given minibatch \(\mathcal{B} = \{1, \ldots, 2n\}\), consisting of \(n\) pairs of augmented samples, the contrastive loss in Equation~\eqref{eq:ntxent} optimizes the alignment of positive pairs \((z_i, z_{i^+})\), while repelling all other samples \(\mathcal{N}(i) = \mathcal{B} \setminus \{i, i^+\}\) as negatives. In general, the negatives \(\mathcal{N}(i)\) can be derived from any combination of embeddings in \(\mathcal{B}\), depending on the composition of the training data.

In self-supervised settings, the training distribution \(P_{\textup{in}}\) typically comprises semantically diverse samples with varied underlying structures, often spanning multiple classes.
This diversity ensures that a minibatch \(\mathcal{B}\) contains sufficiently distinct samples, reducing the likelihood of class collision.

In contrast, anomaly detection tasks deal with \(P_{\textup{in}}\) as a single-class distribution, representing the normal class. These settings often involve highly semantically homogeneous distributions, as seen in industrial defect detection, where the \ac{ID} data contains nearly identical characteristics. Regardless of the degree of homogeneity, in anomaly detection a minibatch \(\mathcal{B}\) is composed exclusively of embeddings derived from \(\mathcal{D}_{\textup{in}}\), the set of \ac{ID} samples drawn from \(P_{\textup{in}}\). The set of possible negatives then becomes:
\[
\mathcal{N}(i) = \{z_a \mid x_a \in \mathcal{D}_{\textup{in}}, a \neq i, i^+\}.
\]
Note that for all \(x_i \in \mathcal{D}_{\textup{in}}\), the corresponding labels satisfy \(y_i = 1\), signifying that all samples in \(\mathcal{D}_{\textup{in}}\) are associated with the normal class.

This lack of semantic diversity among negatives, particularly in more homogeneous cases, leads to exacerbated \textit{class collision}, where semantically similar samples are treated as negatives, forcing embeddings with similar semantics to repel each other. As a result, intra-class variance increases, undermining the goal of learning compact clusters of \ac{ID} representations.

\subsection{Synthetic Outliers as Hard Negatives}\label{sec:synthetic_outliers}
Introducing hard negatives~\citep{robinson2020contrastive} into the training process can significantly improve the quality of the learned representation space in contrastive learning to address the exacerbated class collision issue in anomaly detection. Hard negatives are data points that lie near the boundary of the \ac{ID} distribution \(P_{\textup{in}}\) but are semantically distinct from \ac{ID} samples, providing a stronger contrast during training. In practice, real hard negatives (i.e., actual anomalies) are often unavailable during training in anomaly detection tasks. To overcome this limitation, synthetic outliers are introduced as a surrogate. 

Synthetic outliers are carefully designed to approximate low-density regions near the support boundary of \(P_{\textup{in}}\) offering a challenging contrast for \ac{ID} samples. Formally, let \(\mathcal{D}_{\textup{sout}} = \{x_j\}_{j=1}^M\) represent the set of synthetic outliers, where each \(x_j\) is generated such that:
\[
x_j \notin P_{\textup{in}}, \quad \text{but } x_j \approx \text{Boundary}(P_{\textup{in}}),
\]
where \(\text{Boundary}(P_{\textup{in}})\) refers to the low-density regions at the edge of the \ac{ID}. 

Several methods have been proposed to generate synthetic outliers \(\mathcal{D}_{\textup{sout}}\), including:  
(a) \textit{geometric transformations}, such as those used in RotNet~\citep{golan2018deep}, which apply geometric transformations (e.g., \(90^\circ, 180^\circ, 270^\circ\) rotations) to \ac{ID} samples;  
(b) \textit{local perturbation methods}, such as CutPaste~\citep{li2021cutpaste} and NSA~\citep{schluter2022natural}, which introduce localized image perturbations by randomly cutting patches from an image and pasting them onto other regions;  
(c) \textit{external outlier datasets}, referred to as \ac{OE}~\citep{outlier_exposure}, which use external datasets unrelated to \(P_{\textup{in}}\) to source synthetic outliers \(x_j \in \mathcal{D}_{\textup{sout}}\), ensuring that \(x_j \notin P_{\textup{in}}\), \(x_j \sim P_{\textup{out}}\), and \(P_{\textup{out}} \cap P_{\textup{in}} = \emptyset\); and  
(d) \textit{generative outlier synthesis}, such as DREAM-OOD~\citep{du2024dream} and SIA~\citep{zhang2024realnet}. 

Note that, while lightweight augmentations like rotations or patch perturbations can be effective, the choice of synthetic outlier generator is often domain-specific and can strongly influence performance, making its selection a non-trivial challenge in anomaly detection.

\section{Focused ID Representation Modeling}\label{sec:multi_positive}

\paragraph{Encouraging ID Compactness}
As discussed in Section~\ref{sec:class_collision}, NT-Xent inherently increases intra-class variance by treating semantically similar \ac{ID} samples as negatives. To address this, a natural approach is to expand the set of positives for \ac{ID} samples beyond the single-positive formulation 
\begin{equation}\label{eq:single_positive}
    \Pos(i) = \{i^+\},
\end{equation}
as done in SupCon~\citep{supcon,supcon2}.  
Instead of restricting the positive set to a single augmented view, an effective approach expands it to include all inliers in the batch, while separating them from outliers
\begin{equation}\label{eq:multiple_positives}
    \Pos(i) = \{ j \in \mathcal{B} \mid y_j = y_i \} \setminus \{i\}.
\end{equation}

Given this redefinition of \(\Pos(i)\) in Equation~\eqref{eq:multiple_positives}, each \ac{ID} sample \( i \) now considers all other \ac{ID} samples in \(\mathcal{B}\) as positives, i.e., 
$ \Pos(i) = \{ j \in \mathcal{B} \mid y_j = 1 \} \setminus \{i\}. $
Similarly, for any synthetic outlier \( j \), the positive set includes all synthetic outliers:
$\Pos(j) = \{ k \in \mathcal{B} \mid y_k = y_{\text{sout}} \} \setminus \{j\},$ where $y_{\text{sout}} \neq 1$.

\begin{figure*}
    \scriptsize
    \centering
    \setlength{\tabcolsep}{0pt}
    \includegraphics[width=0.75\textwidth]{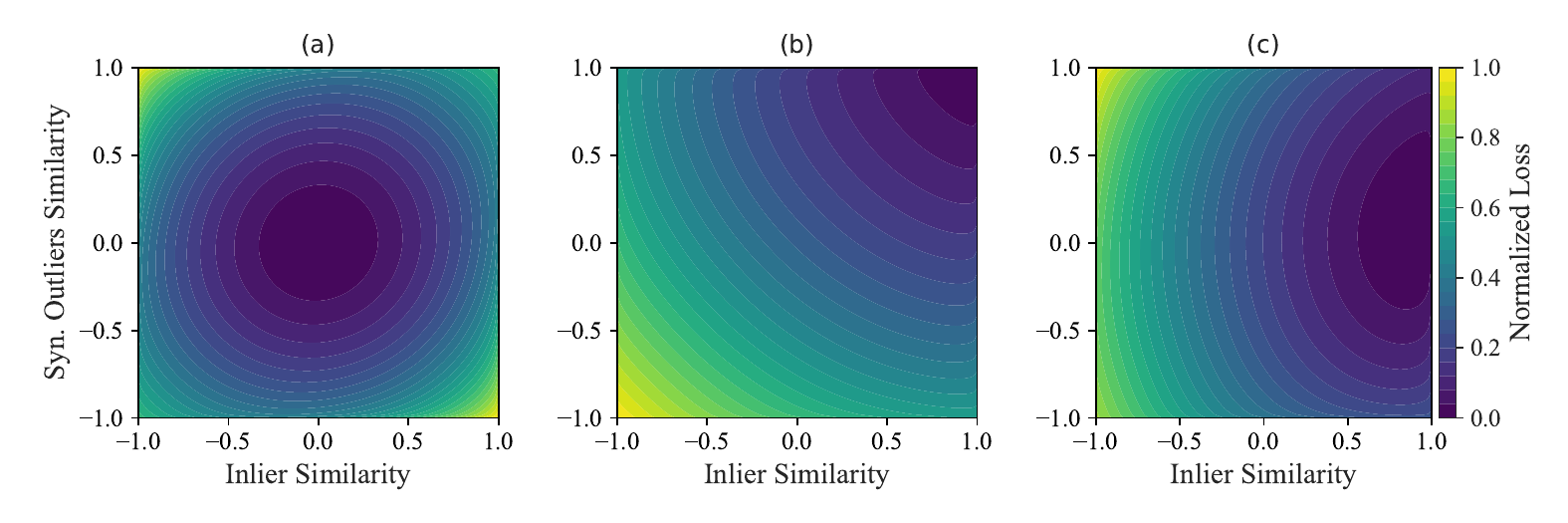}
    \caption{Loss landscape of the contrastive objective in Equation~\eqref{eq:FIRM} under varying definitions of the positive set. In plots (a), (b), and (c), the set $\Pos(i)$ is given by Equations~\eqref{eq:single_positive}, \eqref{eq:multiple_positives}, and \eqref{eq:firm_positives}, respectively.}
\label{fig:loss_functions} 
\end{figure*}

While this expanded definition of \(\Pos(i)\) strengthens \ac{ID} compactness, it overlooks the substantial semantic diversity among anomalies and synthetic outliers. This is particularly evident in real-world anomaly detection tasks, like industrial defect detection, where anomalies can take various forms, as shown in Figure~\ref{fig:mvtec_anomalies}. Similarly, the synthetic outlier generation methods described in Section~\ref{sec:synthetic_outliers} also produce samples with substantial variability. 
Treating synthetic outliers as interchangeable positives disregards their variability and risks excessive alignment among semantically distinct outliers, which, as shown by~\citet{wu2024rethinking}, can negatively impact the quality of learned representations by weakening the training signal when semantically dissimilar positives are aligned.

To investigate this effect, we conduct experiments in semantic anomaly detection settings (i.e., one-class classification), where each dataset class is treated as normal, and synthetic outliers are generated via geometric transformations. Specifically, for a given normal class, we apply rotations of \(90^\circ\), \(180^\circ\), and \(270^\circ\) to produce three distinct synthetic outlier groups. 
We evaluate two learning strategies, differing in how they define the positive set \(\Pos(i)\) for synthetic outliers. Following the positive set definition given by Equation~\eqref{eq:multiple_positives}, the first strategy assigns a shared label space \(\mathcal{Y} \in \{1, 2\}\), where \(y = 1\) represents normal samples and \(y = 2\) denotes all synthetic outliers, treating them as a single contrastive class. In contrast, the second strategy refines the label space to \(\mathcal{Y} \in \{1, 2, 3, 4\}\), where each synthetic outlier group is assigned a distinct label corresponding to its transformation. In this case, synthetic outliers with different transformation types are treated as negatives, enforcing separation between distinct geometric transformations. By encouraging this, the model mitigates representation collapse and maintains diversity in the learned representations, ultimately improving anomaly detection performance. The effect of these two strategies is illustrated in Figure~\ref{fig:supcon_comparison}.

These findings highlight the challenge of balancing \ac{ID} compactness with synthetic outlier separation. An optimal contrastive objective for anomaly detection should satisfy the following key properties:
\begin{enumerate}
    \item[(1)] \textit{Inlier Compactness}: promote the formation of well-clustered representations for \ac{ID} samples, reducing their intra-class variance.
    \item[(2)] \textit{Inlier-Outlier Separation}: enforce inter-class separation between \ac{ID} and synthetic outliers, ensuring that anomalies are projected into distinct, low-density regions of the representation space.
    \item[(3)] \textit{Outlier-Outlier Separation}: maintain sufficient separation among synthetic outliers, preserving diversity to ensure effective contrastive signals.
\end{enumerate}

\subsection{Inter- and Intra-Outlier Separation}

While the properties of inlier compactness and inlier-outlier separation are addressed by including synthetic outliers and expanding the positive set as in Equation~\eqref{eq:multiple_positives}, maintaining separation among synthetic outliers requires a more selective definition of \(\Pos(i)\). To achieve this, we propose a formulation that encourages compact clusters for inliers while preventing collapse among diverse synthetic outliers:
\begin{equation}\label{eq:firm_positives}
    \Pos(i) = \{i^+\} \cup \{j \in \mathcal{B} \setminus \{i\} \mid y_j = 1\}.
\end{equation}

The first term aligns each sample with its augmented counterpart. The second term includes all other inlier samples in the batch and applies only when \(y_i = 1\), encouraging consistent clustering of \ac{ID} representations across the minibatch. For inlier samples \(i \in \mathcal{D}_{\textup{in}}\), where \(y_i = 1\), the second term contributes all other inlier embeddings as positives, reinforcing intra-class consistency. For synthetic outliers \(i \in \mathcal{D}_{\textup{sout}}\), we have \(y_i \neq 1\) by definition, so the second term evaluates to the empty set:
\[
\{j \in \mathcal{B} \setminus \{i\} \mid y_j = 1\} = \emptyset,
\]
which reduces the positive set to \(\Pos(i) = \{i^+\}\). In this case, each synthetic outlier is only aligned with its own augmented view, ensuring that different outliers are treated as negatives to one another. 

\subsection{Training Objective}
With the redefined positive set $\Pos(i)$ given by Equation~\eqref{eq:firm_positives}, we optimize the following contrastive loss to explicitly encourage compact \ac{ID} clustering, separation from synthetic outliers, and separation among them:
\begin{multline}\label{eq:FIRM}
    \mathcal{L}_{\textup{FIRM}}(\mathcal{B}) = 
    \sum_{i \in \mathcal{B}} \frac{1}{|\Pos(i)|} 
    \sum_{p \in \Pos(i)} \\
    -\log \frac{\exp(z_i \bigcdot z_p / \tau)}
    {\sum_{a \in \mathcal{B} \setminus \{i\}} \exp(z_i \bigcdot z_a / \tau)},
\end{multline}
where \(z_i\) represents the normalized embedding of sample \(i\) from the projection network \(g_\psi\), and \(\tau > 0\) is the temperature parameter.

Following the definition of FIRM in Equation~\eqref{eq:FIRM}, we analyze how different formulations of the positive set influence the representation space by visualizing their corresponding loss landscapes. Figure~\ref{fig:loss_functions} illustrates the impact of three different positive set definitions: (a) the single-positive setting given by Equation~\eqref{eq:single_positive}, (b) an expanded positive set that groups all samples of the same class together, given by Equation~\eqref{eq:multiple_positives}, and (c) our proposed formulation in Equation~\eqref{eq:firm_positives}.

\subsection{Detection Score}\label{sec:scores}
Following~\citet{reiss2023mean}, we employ the cosine similarity to the $k$-nearest neighbors as our primary detection score, denoted as:
\begin{equation}\label{eq:score}
s_{\textup{con}}(x, \{x_m\}, k) = \sum_{j \in N_k(x, \{x_m\})} \tilde{f}_{\theta}(x) \bigcdot \tilde{f}_{\theta}(x_j),
\end{equation}
where \(N_k(x, \{x_m\})\) represents the set of $k$-nearest neighbors of the sample $x$ within the set $\{x_m\}$, and \(\tilde{f}_{\theta}(x) = f_{\theta}(x) / \Vert f_{\theta}(x) \Vert\) is the normalized feature embedding of $x$. Note that scoring $x$ involves extracting the representations only for $x$, given that the representations of the set $\{x_m\}$ can be precomputed and stored beforehand.
 
\paragraph{Ensemble scores}
Similarly to \citet{csi}, we employ ensemble scores by incorporating transformations and augmentations during inference. Specifically, the shifting transformation score, \(s_{\textup{shift}}\), averages the cosine similarity across inputs rotated by \(0^\circ, 90^\circ, 180^\circ,\) and \(270^\circ\) degrees. The ensemble score, \(s_{\textup{ens}}\), extends \(s_{\textup{shift}}\) by including multiple random crops for each rotation degree. Detailed formulations of these scores are presented in Appendix~\ref{app:score_functions}.

\section{Patch-Based Learning for Industrial Anomalies}\label{sec:patch_learning}

We adopt the synthetic anomaly generation method from~\citep{zhang2024realnet}, which simulates realistic anomalies by injecting spatial perturbations via strength-controllable diffusion processes. This process relies on two binary masks: (i) a \emph{target foreground mask} \( m \in \{0,1\}^{H \times W} \), obtained via class-specific thresholding and morphological filtering, and (ii) a \emph{synthetic anomaly mask} \( m_{\text{anomaly}} \in \{0,1\}^{H \times W} \), defined as the intersection of \( m \) with a binarized Perlin noise field. We use these masks to localize anomaly injection and constrain spatial sampling during training. 

Unlike prior works, we omit pixel-level supervision and discriminator-based refinement. While such techniques can substantially boost performance, our goal is to isolate the contribution of the proposed loss. We only use the masks to constrain both anomaly injection and positive pair sampling to object regions. Moreover, different from previous works~\citep{yi2020psvdd,defard2021padim,roth2022towards}, where patches are mid-level feature maps, our method defines patches sampled in pixel space.

\paragraph{Foreground-Aware Positive Sampling}
To ensure that contrastive representations capture semantically meaningful object features, we constrain view sampling using the masks \( m \) and \( m_{\text{anomaly}} \). Specifically, we extract two fixed-size square crops, or \emph{patches}, \( \tilde{x}_i, \tilde{x}_{i^+} \in \mathbb{R}^{P \times P \times 3} \), where \( P \) denotes the patch resolution. These patches serve as the augmented instances used in contrastive training and are sampled from the full image \( x \in \mathbb{R}^{H \times W \times 3} \), where \( P < H, W \). 

For \emph{normal} samples, patches are selected such that: (i) the first patch \( \tilde{x}_i \subset x \) must contain pixels such that the fraction of the foreground mask it covers is at least \( \tau_{\text{fore}} \), i.e., \( \frac{|\tilde{x}_i \cap m|}{|m|} \geq \tau_{\text{fore}} \); and (ii) the second patch \( \tilde{x}_{i^+} \subset x \) overlaps with \( \tilde{x}_i \) by at least \(\tau_{\text{over}}\) in image space, i.e., \( \frac{|\tilde{x}_i \cap \tilde{x}_{i^+}|}{|\tilde{x}_i|} \geq \tau_{\text{over}} \). For \emph{synthetic outliers}, the anomaly mask \( m_{\text{anomaly}} \) replaces \( m \), and patches are selected such that: (i) they overlap with each other by at least \(\tau_{\text{over}}\), and (ii) at least one of the two intersects with the anomaly region, i.e., \( |\tilde{x}_j \cap m_{\text{anomaly}}| > 0 \) for some \( j \in \{i, i^+\} \). These thresholds are fixed in our default setting (\(\tau_{\text{fore}} = 0.9\), \(\tau_{\text{over}} = 0.15\)), though tuning them may improve performance in specific applications.

\subsection{Patch-Level Anomaly Scoring and Localization}\label{sec:patch_scoring}
At inference, each image \( x \) is partitioned into overlapping patches of size \( P \times P \) using stride \( S < P \). Let \( \text{patches}(x) \) denote the set of all overlapping patches of size \( P \times P \) extracted from the image \( x \) using stride \( S \). A memory set \( \{p_m\} \) is formed from all patches of normal training images. Using the similarity score \( s_{\textup{con}}(p, \{p_m\}, k) \) from Section~\ref{sec:scores}, with \( k{=}1 \), we define the image-level score as:
\[
s_{\textup{image}}(x) = \max_{p \in \text{patches}(x)} s_{\textup{con}}(p, \{p_m\}, 1).
\]
For localization, patch scores are reshaped into a grid matching the patch layout, upsampled via bilinear interpolation, and smoothed with a Gaussian filter (kernel size 15, \( \sigma = 4 \))~\citep{roth2022towards}.

\begin{table*}[!t]
    \centering
    \scriptsize
    \setlength{\tabcolsep}{0pt}
    \caption{Comparison of contrastive strategies for semantic anomaly detection using different positive set definitions. Mean AUROC (\%) and AULC are reported across all classes on standard benchmarks to reflect performance and convergence.}\label{tab:ablation_objectives}
    \begin{tabularx}{\textwidth}{@{}>{\hsize=2.25cm}L CCCCCCCC@{}}
    \toprule
    \multirow{2}{*}{\textbf{Loss}} & \multicolumn{2}{c}{\textbf{CIFAR-10}} & \multicolumn{2}{c}{\textbf{CIFAR-100 (superclass)}} & \multicolumn{2}{c}{\textbf{FashionMNIST}} & \multicolumn{2}{c}{\textbf{Cats-vs-Dogs}} \\
    \cmidrule(lr){2-3} \cmidrule(lr){4-5} \cmidrule(lr){6-7} \cmidrule(lr){8-9}
     & \textbf{AUROC} & \textbf{AULC} & \textbf{AUROC} & \textbf{AULC} & \textbf{AUROC} & \textbf{AULC} & \textbf{AUROC} & \textbf{AULC} \\  
    \midrule   
    $\NTXent$ & 92.2\sd{0.2} & 89.2\sd{0.3} & 86.3\sd{0.2} & 81.6\sd{0.2} & 95.7\sd{0.1} & 93.7\sd{0.1} & 88.1\sd{0.5} & 81.4\sd{0.4}  \\ 
    SupCon  & 86.5\sd{0.2}  & 64.3\sd{2.7} & {80.7\sd{0.9}} & {60.5\sd{2.1}} & 96.4\sd{0.1} & 87.5\sd{4.7} & 58.2\sd{0.1} & 51.5\sd{0.5}  \\ 
    Rot-SupCon & 92.5\sd{0.2}  & 90.5\sd{0.1} & 85.9\sd{0.4} & 82.5\sd{0.3} & 96.6\sd{0.1} & 95.5\sd{0.1} & 89.2\sd{0.1} & 85.3\sd{0.2} \\ 
    $\FIRM$     & \textbf{93.4\sd{0.2}} & \textbf{91.9\sd{0.1}} & \textbf{87.8\sd{0.2}} & \textbf{85.1\sd{0.2}} & \textbf{96.8\sd{0.1}} & \textbf{95.7\sd{0.1}} & \textbf{90.4\sd{0.4}} & \textbf{87.8\sd{0.6}}  \\ 
    \bottomrule
    \end{tabularx}
\end{table*}  
 
\begin{figure*}[!t]
    \scriptsize
    \centering
    \setlength{\tabcolsep}{0pt}
    \begin{tabular}{cccc} \includegraphics[width=0.25\textwidth]{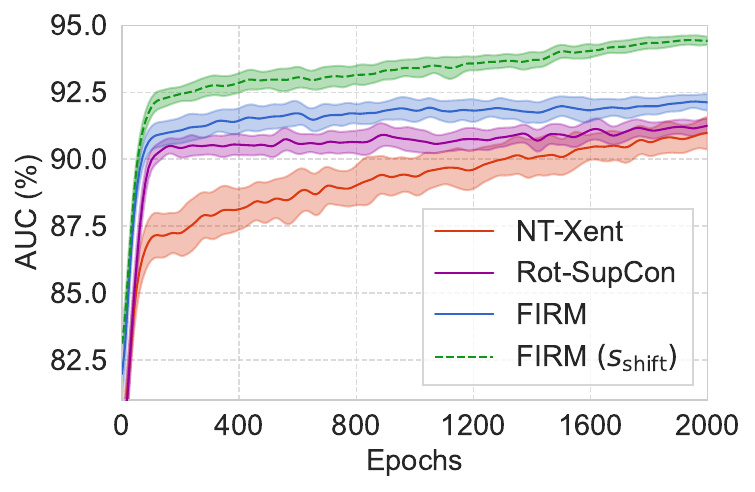} &  \includegraphics[width=0.25\textwidth]{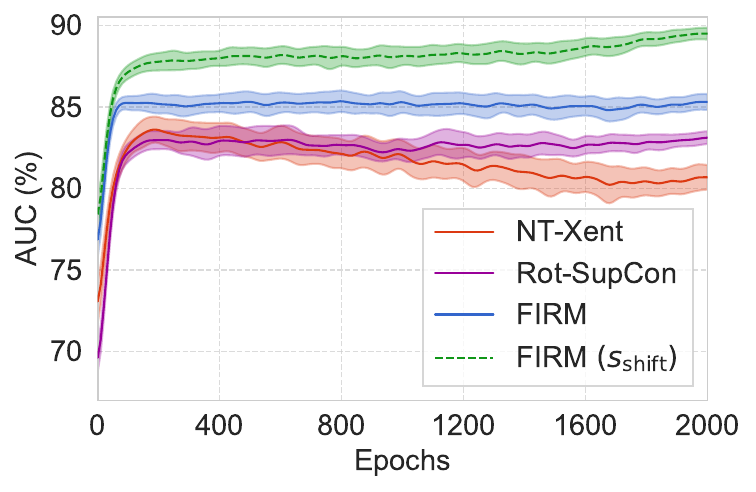} &  \includegraphics[width=0.25\textwidth]{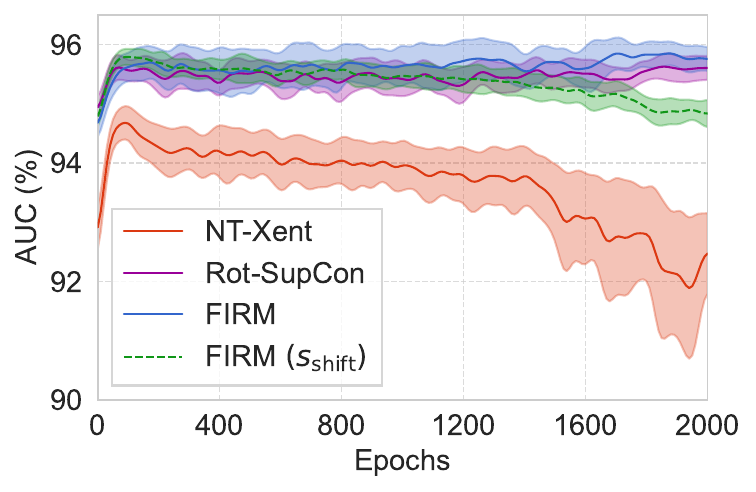} &  \includegraphics[width=0.25\textwidth]{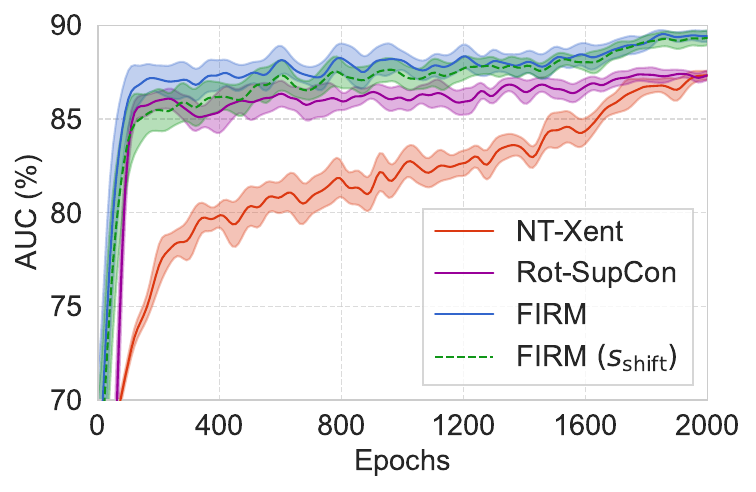}  \\
        (a) CIFAR-10. & (b) CIFAR-100 (superclass). & (c) Fashion-MNIST.  & (d) Cats-vs-Dogs.
    \end{tabular}
    \caption{Learning curves of different contrastive strategies over 2000 training epochs. Each plot shows the mean AUROC and its variance. FIRM demonstrates faster convergence and more stable performance across epochs.}\label{fig:learning_curves}
\end{figure*}

\section{Experiments}\label{sec:experiments}

This section presents the experimental results evaluating our proposed objective on \textit{semantic anomaly detection} and \textit{industrial anomaly detection}. Our code for the \ac{FIRM} loss can be found at \url{https://github.com/willtl/firm}.

\paragraph{Evaluation Scope} In addition to industrial anomaly experiments, we conduct extensive evaluations on semantic anomaly detection benchmarks to systematically analyze the behavior of contrastive learning objectives under varying anomaly types, dataset distributions, and outlier generation strategies. These datasets offer a controlled setting 
in which normal and anomalous classes are explicitly defined, and where synthetic outliers can be consistently generated and interpreted. The semantic anomaly detection experiments follow the settings defined in~\citep{golan2018deep,hendrycks2019using,bergman2020classification}, using benchmarks including CIFAR-10/CIFAR-100~\citep{cifar100_dataset}, Fashion-MNIST~\citep{fmnist_dataset}, and Cats-vs-Dogs~\citep{catvsdog_dataset}, where each class is considered normal while the others are treated as anomalies. For CIFAR-100, we use the superclass setting to create less homogeneous \ac{ID} classes. The industrial anomaly detection experiments use the MVTec-AD dataset~\citep{bergmann2021mvtec}, where normal samples exhibit structural homogeneity, and anomalies manifest as subtle defects. We use the official training and test splits provided by each dataset.
For additional experiments, refer to Appendix~\ref{app:ablation_studies}. 

\paragraph{Experimental Setup} All experiments use the ResNet-18 architecture~\citep{he2016deep}. Synthetic outliers are generated using three approaches. For semantic anomaly detection, we apply deterministic transformations \( \Omega_\gamma: \mathcal{X} \to \mathcal{X} \) with \( \gamma \in \{90^\circ, 180^\circ, 270^\circ\} \) to inliers from \( \mathcal{D}_{\text{in}} \). 
For industrial anomaly detection, we use NSA~\citep{schluter2022natural} during ablations, which generates natural synthetic defects through localized patch-based perturbations with Poisson blending. For the main results, we adopt SIA~\citep{zhang2024realnet}, which leverages a strength-controllable diffusion process to synthesize more realistic and diverse anomalies, thereby improving alignment with real-world defect distributions.
For \ac{OE}, we use a curated subset of the 80 Million Tiny Images dataset, specifically the 300,000-image set from~\citep{outlier_exposure}, debiased to remove overlaps. Experiments with \ac{OE} are referred to as ``FIRM + OE.'' We report the mean and standard deviation over five runs. CIFAR-10, CIFAR-100, Fashion-MNIST, and Cats-vs-Dogs images are resized to 32x32 and 64x64 pixels, respectively~\citep{learn_and_eval}, and MVTec images are resized to 256x256 pixels~\citep{schluter2022natural}. For MVTec experiments using the patch-based strategy, class-dependent patch sizes ranging from \(16 \times 16\) to \(64 \times 64\) are used during training, while fixed patches of \(32 \times 32\) with a stride of 16 are extracted during evaluation. Further details are in Appendix~\ref{app:training_details}.

\paragraph{Rot-SupCon Definition}
For the methods used in our ablation experiments, we reiterate the definition of the positive set $\Pos(i)$. The positive set definitions for NT-Xent, SupCon, and FIRM are given by Equations~\eqref{eq:single_positive}, \eqref{eq:multiple_positives}, and \eqref{eq:firm_positives}, respectively. Rot-SupCon refines SupCon's positive set by labeling synthetic outliers based on transformation types (\(\mathcal{Y} \in \{1,2,3,4\}\)). Note Rot-SupCon is unsuitable for scenarios where outliers lack a clear label structure, such as in the industrial synthetic anomaly generation process or with unlabeled external data for \ac{OE}.

\begin{table}[!t]
    \centering
    \scriptsize
    \setlength{\tabcolsep}{0pt}  
    \caption{Ablation study on MVTec-AD using full-image training with image-level AUROC (\%). Unlike the main results (Table~\ref{tab:main_industrial_results}), this setup uses NSA for synthetic anomalies and trains directly on whole images, isolating the contribution of the SIA generation process and the patch-based strategy.}\label{tab:results_mvtec_fullwidth} 
    \begin{tabularx}{\linewidth}{@{} L CCC @{}}
        \toprule
        \textbf{Classes} & \textbf{NT-Xent} & \textbf{SupCon} & \textbf{FIRM} \\
        \midrule
        Bottle & \textbf{100\sd{0.0}} & \textbf{100\sd{0.0}} & \textbf{100\sd{0.0}} \\
        Cable & 97.4\sd{0.6} & 77.0\sd{0.6} & \textbf{97.4\sd{0.3}} \\
        Capsule & 80.7\sd{1.6} & 73.6\sd{0.5} & \textbf{92.8\sd{0.2}} \\
        Hazelnut & 84.9\sd{4.3} & 95.7\sd{0.7} & \textbf{96.5\sd{0.1}} \\
        Metal Nut & 95.9\sd{0.4} & 80.8\sd{2.7} & \textbf{98.6\sd{0.2}} \\
        Pill & 90.0\sd{0.1} & \textbf{97.0\sd{0.0}} & 92.8\sd{0.6} \\
        Screw & 72.7\sd{10.7} & 39.8\sd{6.4} & \textbf{96.7\sd{0.5}} \\
        Toothbrush & 98.6\sd{0.3} & \textbf{100\sd{0.0}} & \textbf{100\sd{0.0}} \\
        Transistor & \textbf{94.3\sd{1.0}} & 81.8\sd{9.3} & 93.1\sd{0.3} \\
        Zipper & 99.8\sd{0.2} & \textbf{100\sd{0.0}} & \textbf{100\sd{0.0}} \\
        \midrule
        Carpet & 65.4\sd{6.2} & 67.6\sd{8.0} & \textbf{75.2\sd{0.3}} \\
        Grid & 94.6\sd{0.8} & 85.5\sd{1.0} & \textbf{100\sd{0.0}} \\
        Leather & 88.4\sd{0.8} & 86.0\sd{0.4} & \textbf{93.8\sd{0.8}} \\
        Tile & 99.8\sd{0.1} & 83.6\sd{1.8} & \textbf{100\sd{0.0}} \\
        Wood & 87.2\sd{4.7} & \textbf{96.8\sd{1.1}} & 87.5\sd{0.3} \\
        \midrule
        \textbf{Mean} & 90.0\sd{2.1} & 84.3\sd{2.2} & \textbf{95.0\sd{0.2}} \\
        \bottomrule
    \end{tabularx}
\end{table}

\begin{table*}[!t]
    \centering
    \scriptsize
    \setlength{\tabcolsep}{4pt}
    \caption{AUROC (\%) results on the MVTec-AD dataset. Values are reported in the format \textit{image-level / pixel-level}.}
    \label{tab:main_industrial_results}
    \begin{tabularx}{\linewidth}{@{} c >{\hsize=1.35cm}X >{\hsize=1.3cm}C CC>{\hsize=1.3cm}CC>{\hsize=1.3cm}C>{\hsize=1.3cm}CC @{}}
        \toprule
        \textbf{Group} & \textbf{Category} & \textbf{P-SVDD} & \textbf{CutPaste} & \textbf{NSA} & \textbf{DRAEM} & \textbf{DeSTSeg} & \textbf{DiffAD} & \textbf{PatchCore} & \textbf{FIRM} \\
        \midrule
        \multirow{10}{*}{\rotatebox[origin=c]{90}{Object}} & Bottle & 98.6 / 98.1 & 98.3\sd{0.5} / 97.6\sd{0.1} & 97.7\sd{0.3} / 93.8\sd{0.1} & 99.2 / 99.1 & -- / \textbf{99.2\sd{0.2}} & \textbf{100} / 98.8 & \textbf{100} / 98.6 & \textbf{100\sd{0.0}} / 99.1\sd{0.0} \\
         & Cable & 90.3 / 96.8 & 80.6\sd{0.5} / 90.0\sd{0.2} & 94.5\sd{1.0} / 96.0\sd{1.4} & 91.8 / 94.7 & -- / 97.3\sd{0.4} & 94.6 / 96.8 & \textbf{99.5} / \textbf{98.4} & 98.3\sd{0.2} / 96.3\sd{0.1} \\
         & Capsule & 76.7 / 95.8 & 96.2\sd{0.5} / 97.4\sd{0.1} & 95.2\sd{1.7} / 97.6\sd{0.9} & 98.5 / 94.3 & -- / \textbf{99.1\sd{0.0}} & 97.5 / 98.2 & 98.1 / 98.8 & \textbf{98.6\sd{0.2}} / 98.0\sd{0.2} \\
         & Hazelnut & 92.0 / 97.5 & 97.3\sd{0.3} / 97.3\sd{0.1} & 94.7\sd{1.1} / 97.6\sd{0.6} & \textbf{100} / \textbf{99.7} & -- / 96.1\sd{2.2} & \textbf{100} / 99.4 & \textbf{100} / 99.3 & 98.7\sd{0.4} / 98.5\sd{0.0} \\
         & Metal Nut & 94.0 / 98.0 & 99.3\sd{0.2} / 93.1\sd{0.4} & 98.7\sd{0.7} / 98.4\sd{0.2} & 98.7 / \textbf{99.5} & -- / 99.1\sd{0.1} & 99.5 / 99.1 & 96.6 / 98.4 & \textbf{100\sd{0.0}} / 98.4\sd{0.0} \\
         & Pill & 86.1 / 95.1 & 92.4\sd{1.3} / 95.7\sd{0.1} & \textbf{99.2\sd{0.6}} / 98.5\sd{0.3} & 98.9 / 97.6 & -- / \textbf{99.6\sd{0.2}} & 97.7 / 97.7 & 98.1 / 97.4 & 96.7\sd{0.4} / 98.8\sd{0.1} \\
         & Screw & 81.3 / 95.7 & 86.3\sd{1.0} / 96.7\sd{0.1} & 90.2\sd{1.4} / 96.5\sd{1.0} & 93.9 / 97.6 & -- / \textbf{99.7\sd{0.0}} & 97.2 / 99.0 & 98.1 / 99.4 & \textbf{99.9\sd{0.7}} / 98.4\sd{0.2} \\
         & Toothbrush & \textbf{100} / 98.1 & 98.3\sd{0.9} / 98.1\sd{0.0} & \textbf{100\sd{0.0}} / 94.7\sd{0.7} & \textbf{100} / 98.1 & -- / 98.6\sd{0.4} & \textbf{100} / \textbf{99.2} & \textbf{100} / 98.7 & \textbf{100\sd{0.0}} / 97.7\sd{0.0} \\
         & Transistor & 91.5 / 97.0 & 95.5\sd{0.5} / 93.0\sd{0.2} & 95.1\sd{0.2} / 94.2\sd{0.3} & 93.1 / 90.9 & -- / \textbf{98.7\sd{0.4}} & 96.1 / 93.7 & \textbf{100} / 96.3 & 98.0\sd{0.1} / 97.4\sd{0.1} \\
         & Zipper & 97.9 / 95.1 & 99.4\sd{0.2} / \textbf{99.3\sd{0.0}} & 99.8\sd{0.1} / \textbf{99.3\sd{0.1}} & \textbf{100} / 98.8 & -- / 98.5\sd{0.3} & \textbf{100} / 99.0 & 99.4 / 98.8 & \textbf{100\sd{0.0}} / 98.7\sd{0.0} \\
        \cmidrule(lr){2-10}
        & \textbf{Object Avg.} & 90.8 / 96.7 & 94.4\sd{0.6} / 95.8\sd{0.1} & 96.5\sd{0.7} / 96.7\sd{0.6} & 97.4 / 97.0 & -- / 98.6\sd{0.4} & 98.3 / 98.1 & \textbf{99.0} / 98.4 & \textbf{99.0\sd{0.2}} / 98.1\sd{0.1} \\
        \midrule
        \multirow{5}{*}{\rotatebox[origin=c]{90}{Texture}} & Carpet & 92.9 / 92.6 & 93.1\sd{1.1} / 98.3\sd{0.0} & 95.6\sd{0.6} / 95.5\sd{2.3} & 97.0 / 95.5 & -- / 98.0\sd{0.7} & 98.3 / 98.1 & \textbf{98.7} / \textbf{99.0} & 98.5\sd{0.8} / 98.5\sd{0.3} \\
         & Grid & 94.6 / 96.2 & 99.9\sd{0.1} / 97.5\sd{0.1} & 99.9\sd{0.1} / \textbf{99.9\sd{0.1}} & 99.9 / 99.7 & -- / 99.3\sd{0.1} & \textbf{100} / 99.7 & 98.2 / 98.7 & \textbf{100\sd{0.0}} / 96.8\sd{0.2} \\
         & Leather & 90.9 / 97.4 & \textbf{100\sd{0.0}} / \textbf{99.5\sd{0.0}} & 99.9\sd{0.1} / \textbf{99.5\sd{0.1}} & \textbf{100} / 98.6 & -- / 89.1\sd{3.4} & \textbf{100} / 99.1 & \textbf{100} / 99.3 & \textbf{100\sd{0.0}} / 99.3\sd{0.0} \\
         & Tile & 97.8 / 91.4 & 93.4\sd{1.0} / 90.5\sd{0.2} & \textbf{100\sd{0.0}} / 99.3\sd{0.0} & 99.6 / 99.2 & -- / 97.7\sd{0.3} & \textbf{100} / \textbf{99.4} & 98.7 / 95.6 & \textbf{100\sd{0.0}} / 99.2\sd{0.1} \\
         & Wood & 96.5 / 90.8 & 98.6\sd{0.5} / 95.5\sd{0.1} & 97.5\sd{1.5} / 90.7\sd{1.9} & 99.1 / 96.4 & -- / \textbf{99.1\sd{0.5}} & \textbf{100} / 96.7 & 99.2 / 95.0 & \textbf{100\sd{0.0}} / 97.1\sd{0.1} \\
        \cmidrule(lr){2-10}
        & \textbf{Texture Avg.} & 94.5 / 93.7 & 97.0\sd{0.5} / 96.3\sd{0.1} & 98.6\sd{0.5} / 97.0\sd{0.9} & 99.1 / 97.9 & -- / 96.6\sd{1.0} & \textbf{99.7} / 98.6 & 99.0 / 97.5 & \textbf{99.7\sd{0.2}} / 98.2\sd{0.1} \\
        \midrule
        \multicolumn{2}{l}{\textbf{Overall Avg.}} & 92.1 / 95.7 & 95.2\sd{0.6} / 96.0\sd{0.1} & 97.2\sd{0.6} / 96.8\sd{0.7} & 98.0 / 97.3 & 98.6 / 97.9\sd{0.6} & 98.7 / \textbf{98.3} & 99.0 / 98.1 & \textbf{99.2\sd{0.2}} / 98.1\sd{0.1} \\
        \bottomrule
    \end{tabularx}
\end{table*}
 
\begin{table}[t]
\centering
\scriptsize
\caption{Semantic anomaly detection results in AUROC (\%). Full per-class and additional ablations are reported in Appendix B.}
\label{tab:main_semantic_results}
\begin{tabularx}{\columnwidth}{@{}l*{4}{>{\centering\arraybackslash}X}@{}}
\toprule
\textbf{Method}          & \textbf{CIFAR-10}
                         & \textbf{CIFAR-100}
                         & \textbf{F-MNIST}
                         & \textbf{Cats vs Dogs} \\
\midrule
GOAD                     & 88.2               & --                  & 94.1 & --               \\
SSD                      & 90.0               & --                  & --            & --               \\
Rot.\ Pred.              & 91.3\sd{0.3}                  & 84.1\sd{0.6}       & 95.8\sd{0.3} & 86.4\sd{0.6} \\
DROC                     & 92.5\sd{0.6}               & 86.5\sd{0.7}       & 94.5\sd{0.4} & 89.6\sd{0.3} \\
CSI                      & 94.3               & 89.6               & --            & --            \\
\midrule
FIRM                    & \textbf{95.4\sd{0.1}}      & \textbf{91.0\sd{0.2}} & \textbf{96.8\sd{0.1}} & \textbf{90.4\sd{0.5}} \\
FIRM + OE               & \textbf{97.6\sd{0.0}}               & \textbf{94.7\sd{0.1}}       & \textbf{96.2\sd{0.1}} & --            \\
\bottomrule
\end{tabularx}
\vspace{-0.4em}
\end{table}

\subsection{Ablation Studies}\label{sec:ablation}
We conduct ablation studies to evaluate the impact of different positive set definitions on anomaly detection performance in both semantic and industrial settings, focusing on detection scores and synthetic outlier sources. Additional ablation studies on batch size, temperature, and scoring methods are presented in Appendix~\ref{app:ablation_studies}.

\paragraph{Semantic anomaly detection}  
We compare FIRM with NT-Xent, SupCon, and Rot-SupCon under semantic anomaly detection settings to analyze how the sampling and internal organization of the positive set influence the learning process. We use cosine similarity $\scon$ with $k = 1$ for all methods to ensure a fair comparison. Performance is measured using the mean \ac{AUROC} and the \ac{AULC}, where \ac{AULC} captures peak performance and convergence efficiency.

Table~\ref{tab:ablation_objectives} shows that FIRM consistently outperforms all other formulations across datasets. Its superior AUROC and AULC suggest that FIRM guides optimization in the parameter space more effectively, leading to representations that better align with the downstream anomaly detection objective. The improved convergence speed and stability indicate that FIRM provides a more structured contrastive signal. Rot-SupCon exhibits strong performance but fails in binary setups (SupCon). This highlights the importance of encouraging separation among synthetic outliers. Such separation is essential for maintaining contrastive signal integrity and preventing representational collapse in low-diversity settings. Figure~\ref{fig:learning_curves} further illustrates the convergence dynamics. While NT-Xent and standard SupCon struggle with unstable \ac{AUROC} across epochs, FIRM reaches peak performance significantly faster, plateauing around 400 epochs for CIFAR-10 and CIFAR-100. 

\paragraph{Industrial anomaly detection}  
To evaluate the generalizability of FIRM to industrial settings, we perform experiments on MVTec-AD using a simplified protocol designed to isolate the effect of the contrastive objective itself. Specifically, we use NSA~\citep{schluter2022natural} to generate synthetic anomalies via localized cut-and-paste perturbations, and we omit the patch-based learning strategy described in Section~\ref{sec:multi_positive}. Instead, we train on full-resolution images, using simple augmentations consisting of small translations (up to 10\%) and light color jitter to generate augmented image pairs. This setting allows for controlled comparisons across contrastive formulations and enables side-by-side evaluation with the full patch-based approach reported in Table~\ref{tab:main_industrial_results}.

As shown in Table~\ref{tab:results_mvtec_fullwidth}, FIRM achieves a mean AUROC of 95.0\%, surpassing NT-Xent by 5.0\% and SupCon by 10.7\%. Results show that FIRM consistently outperforms other methods, particularly on categories involving subtle and fine-grained defects, such as minor texture inconsistencies or small structural anomalies. Unlike Rot-SupCon, which relies on labeled synthetic transformations, FIRM remains effective without requiring explicit outlier categorization, encouraging diversity in a label-agnostic manner. 

\subsection{Main Results}\label{sec:main_results}

Building on the ablation findings, we now present the main experimental results. We assess the effectiveness of FIRM on semantic anomaly detection using three scoring strategies: cosine similarity to nearest neighbors ($\scon$), an ensemble of rotated views ($\sshift$), and a full ensemble combining crops and rotations ($\sens$). These progressively incorporate more spatial and contextual variation to strengthen detection robustness. We also evaluate the variant ``FIRM + OE,'' which includes additional external outliers during training. OE results are excluded for Cats-vs-Dogs due to resolution mismatches in the OE dataset. Full per-class AUROC scores can be found in Appendix~\ref{app:per_class_results}.

For industrial anomaly detection, in contrast to the ablation setup, which relied on NSA-generated synthetic outliers and whole-images, the following results reflect the full experimental setting using patch-based learning and evaluation, as described in Section~\ref{sec:scores}. This setup uses synthetic anomalies generated via SIA~\citep{zhang2024realnet}, a diffusion-based method that produces realistic and fine-grained defects.

\paragraph{Semantic anomaly detection}  
For semantic anomaly detection, we compare FIRM to a range of self-supervised and contrastive methods that do not rely on explicit anomaly labels. These include GOAD~\citep{bergman2020classification}, SSD~\citep{sehwag2021ssd}, Rotational Prediction~\citep{learn_and_eval}, DROC~\citep{learn_and_eval}, and CSI~\citep{csi}. Among these, DROC and CSI are particularly strong baselines as they also employ contrastive learning with rotation-based synthetic outliers. DROC uses the InfoNCE loss. CSI, on the other hand, combines the NT-Xent loss with a cross-entropy objective over rotation labels. This combination is conceptually similar to the Rot-SupCon strategy introduced in Section~\ref{sec:multi_positive}.

Table~\ref{tab:main_semantic_results} presents overall AUROC scores across four semantic anomaly detection benchmarks. FIRM achieves the highest performance across all datasets, with further gains from FIRM+OE. The OE variant introduces only modest data-handling overhead, as architecture, hyperparameters, and training steps match the synthetic-outlier setting; gains stem from richer negative signals rather than increased compute. Full per-class results are in Appendix~\ref{app:per_class_results}, Tables~\ref{tab:cipher10_per_class_classifiers}–\ref{tab:catsdogs_per_class_classifiers}.

\paragraph{Industrial anomaly detection}  
For industrial anomaly detection, we evaluate FIRM on MVTec-AD and compare it to recent state-of-the-art methods, including P-SVDD~\citep{yi2020psvdd}, CutPaste~\citep{li2021cutpaste}, NSA (logistic)~\citep{schluter2022natural}, DRAEM~\citep{zavrtanik2021draem}, PatchCore~\citep{roth2022towards}, DeSTSeg~\citep{zhang2023destseg}, which enhances student-teacher frameworks with denoising and synthetic mask supervision, and DiffAD~\citep{zhang2023unsupervised}, which leverages latent diffusion for robust anomaly reconstruction and localization.

Table~\ref{tab:main_industrial_results} presents AUROC (\%) results on the MVTec-AD dataset, with scores shown in the format \textit{image-level / pixel-level}, where a dash (--) indicates that the corresponding value was not reported by the original method. Both FIRM and PatchCore achieve strong image-level performance, with PatchCore obtaining 99.0\% on object and texture categories, and FIRM achieving 99.0\% and 99.7\% respectively, leading to FIRM's best overall image-level AUROC of 99.2\%. At the pixel level, DeSTSeg achieves the highest score for object categories, while DiffAD obtains the highest performance for textures; overall, DiffAD attains the best mean pixel-level AUROC of 98.3\%. Compared to the simplified NSA-based full-image training setup used in our ablations (Table~\ref{tab:results_mvtec_fullwidth}), the combination of SIA-generated outliers and patch-based training yields substantially stronger results.

FIRM uses a fixed patch stride of 16 during evaluation. While prior work has explored architectural and refinement strategies to boost performance~\citep{zhang2023destseg,zhang2023unsupervised}, our goal here is to isolate the contrastive objective’s effects rather than optimize exhaustively. Further gains may come from stride tuning, receptive field adjustment, or task-specific post-processing. Notably, encouraging test-time patches to overlap more with object regions can improve localization accuracy, suggesting a promising refinement direction. Finally, although FIRM is designed for image-based anomalies, extending it to handle time-varying and evolving anomalies remains an active area of investigation.

\section{Conclusion}
In this work, we introduced FIRM, a contrastive learning framework for anomaly detection that enforces three key properties for effective detection: inlier compactness, inlier–outlier separation, and outlier diversity. By selectively structuring the positive set, FIRM yields robust representations and extends naturally to a patch-based setting for industrial anomaly detection, enabling fine-grained localization without pixel-level supervision. Experiments on semantic and industrial benchmarks show that FIRM outperforms traditional contrastive objectives and recent methods, with superior stability, robustness, and gains from external data. 

\bibliography{mybibfile}

\begin{thebibliography}{46}
\providecommand{\natexlab}[1]{#1}
\providecommand{\url}[1]{\texttt{#1}}
\expandafter\ifx\csname urlstyle\endcsname\relax
  \providecommand{\doi}[1]{doi: #1}\else
  \providecommand{\doi}{doi: \begingroup \urlstyle{rm}\Url}\fi

\bibitem[Bergman and Hoshen(2020)]{bergman2020classification}
L.~Bergman and Y.~Hoshen.
\newblock Classification-based anomaly detection for general data.
\newblock In \emph{International Conference on Learning Representations}, 2020.

\bibitem[Bergmann et~al.(2021)Bergmann, Batzner, Fauser, Sattlegger, and Steger]{bergmann2021mvtec}
P.~Bergmann, K.~Batzner, M.~Fauser, D.~Sattlegger, and C.~Steger.
\newblock The mvtec anomaly detection dataset: a comprehensive real-world dataset for unsupervised anomaly detection.
\newblock \emph{International Journal of Computer Vision}, 129\penalty0 (4):\penalty0 1038--1059, 2021.

\bibitem[Bucci et~al.(2022)Bucci, Borlino, Caputo, and Tommasi]{bucci2022distance}
S.~Bucci, F.~C. Borlino, B.~Caputo, and T.~Tommasi.
\newblock Distance-based hyperspherical classification for multi-source open-set domain adaptation.
\newblock In \emph{Proceedings of the IEEE/CVF Winter Conference on Applications of Computer Vision}, pages 1119--1128, 2022.

\bibitem[Chen et~al.(2020{\natexlab{a}})Chen, Kornblith, Norouzi, and Hinton]{simclr}
T.~Chen, S.~Kornblith, M.~Norouzi, and G.~Hinton.
\newblock A simple framework for contrastive learning of visual representations.
\newblock In \emph{International conference on machine learning}, pages 1597--1607. PMLR, 2020{\natexlab{a}}.

\bibitem[Chen et~al.(2020{\natexlab{b}})Chen, Fan, Girshick, and He]{mocov2}
X.~Chen, H.~Fan, R.~Girshick, and K.~He.
\newblock Improved baselines with momentum contrastive learning.
\newblock \emph{CoRR}, abs/2003.04297, 2020{\natexlab{b}}.

\bibitem[Chen et~al.(2021)Chen, Xie, and He]{mocov3}
X.~Chen, S.~Xie, and K.~He.
\newblock An empirical study of training self-supervised vision transformers.
\newblock In \emph{Proceedings of the IEEE/CVF international conference on computer vision}, pages 9640--9649, 2021.

\bibitem[Defard et~al.(2021)Defard, Setkov, Loesch, and Audigier]{defard2021padim}
T.~Defard, A.~Setkov, A.~Loesch, and R.~Audigier.
\newblock Padim: a patch distribution modeling framework for anomaly detection and localization.
\newblock In \emph{International conference on pattern recognition}, pages 475--489. Springer, 2021.

\bibitem[Du et~al.(2022)Du, Wang, Cai, and Li]{du2022vos}
X.~Du, Z.~Wang, M.~Cai, and Y.~Li.
\newblock Vos: Learning what you don’t know by virtual outlier synthesis.
\newblock \emph{Proceedings of the International Conference on Learning Representations}, 2022.

\bibitem[Du et~al.(2024)Du, Sun, Zhu, and Li]{du2024dream}
X.~Du, Y.~Sun, J.~Zhu, and Y.~Li.
\newblock Dream the impossible: Outlier imagination with diffusion models.
\newblock \emph{Advances in Neural Information Processing Systems}, 36, 2024.

\bibitem[Elson et~al.(2007)Elson, Douceur, Howell, and Saul]{catvsdog_dataset}
J.~Elson, J.~Douceur, J.~Howell, and J.~Saul.
\newblock Asirra: A captcha that exploits interest-aligned manual image categorization.
\newblock In \emph{Proceedings of the 14th ACM Conference on Computer and Communications Security (CCS)}, volume~7, pages 366--374. ACM, 2007.

\bibitem[Ghosal et~al.(2024)Ghosal, Sun, and Li]{ghosal2024overcome}
S.~S. Ghosal, Y.~Sun, and Y.~Li.
\newblock How to overcome curse-of-dimensionality for out-of-distribution detection?
\newblock In \emph{Proceedings of the AAAI Conference on Artificial Intelligence}, volume~38, pages 19849--19857, 2024.

\bibitem[Gidaris et~al.(2018)Gidaris, Singh, and Komodakis]{rotpred}
S.~Gidaris, P.~Singh, and N.~Komodakis.
\newblock Unsupervised representation learning by predicting image rotations.
\newblock In \emph{International Conference on Learning Representations}, 2018.

\bibitem[Golan and El-Yaniv(2018)]{golan2018deep}
I.~Golan and R.~El-Yaniv.
\newblock Deep anomaly detection using geometric transformations.
\newblock \emph{Advances in neural information processing systems}, 31, 2018.

\bibitem[He et~al.(2016)He, Zhang, Ren, and Sun]{he2016deep}
K.~He, X.~Zhang, S.~Ren, and J.~Sun.
\newblock Deep residual learning for image recognition.
\newblock In \emph{2016 IEEE Conference on Computer Vision and Pattern Recognition (CVPR)}, pages 770--778, 2016.

\bibitem[He et~al.(2020)He, Fan, Wu, Xie, and Girshick]{moco}
K.~He, H.~Fan, Y.~Wu, S.~Xie, and R.~Girshick.
\newblock Momentum contrast for unsupervised visual representation learning.
\newblock In \emph{Proceedings of the IEEE/CVF conference on computer vision and pattern recognition}, pages 9729--9738, 2020.

\bibitem[Hendrycks et~al.(2018)Hendrycks, Mazeika, and Dietterich]{outlier_exposure}
D.~Hendrycks, M.~Mazeika, and T.~Dietterich.
\newblock Deep anomaly detection with outlier exposure.
\newblock In \emph{International Conference on Learning Representations}, 2018.

\bibitem[Hendrycks et~al.(2019)Hendrycks, Mazeika, Kadavath, and Song]{hendrycks2019using}
D.~Hendrycks, M.~Mazeika, S.~Kadavath, and D.~Song.
\newblock Using self-supervised learning can improve model robustness and uncertainty.
\newblock \emph{Advances in neural information processing systems}, 32, 2019.

\bibitem[Khosla et~al.(2020)Khosla, Teterwak, Wang, Sarna, Tian, Isola, Maschinot, Liu, and Krishnan]{supcon}
P.~Khosla, P.~Teterwak, C.~Wang, A.~Sarna, Y.~Tian, P.~Isola, A.~Maschinot, C.~Liu, and D.~Krishnan.
\newblock Supervised contrastive learning.
\newblock \emph{Advances in neural information processing systems}, 33:\penalty0 18661--18673, 2020.

\bibitem[Krizhevsky et~al.(2009)]{cifar100_dataset}
A.~Krizhevsky et~al.
\newblock Learning multiple layers of features from tiny images.
\newblock Technical report, University of Toronto, 2009.

\bibitem[Li et~al.(2021)Li, Sohn, Yoon, and Pfister]{li2021cutpaste}
C.-L. Li, K.~Sohn, J.~Yoon, and T.~Pfister.
\newblock Cutpaste: Self-supervised learning for anomaly detection and localization.
\newblock In \emph{Proceedings of the IEEE/CVF conference on computer vision and pattern recognition}, pages 9664--9674, 2021.

\bibitem[Ming et~al.(2023)Ming, Sun, Dia, and Li]{ming2023cider}
Y.~Ming, Y.~Sun, O.~Dia, and Y.~Li.
\newblock How to exploit hyperspherical embeddings for out-of-distribution detection?
\newblock In \emph{The Eleventh International Conference on Learning Representations}, 2023.

\bibitem[Oord et~al.(2018)Oord, Li, and Vinyals]{oord2018representation}
A.~v.~d. Oord, Y.~Li, and O.~Vinyals.
\newblock Representation learning with contrastive predictive coding.
\newblock \emph{CoRR}, abs/1807.03748, 2018.

\bibitem[Parzen(1962)]{parzen1962estimation}
E.~Parzen.
\newblock On estimation of a probability density function and mode.
\newblock \emph{The annals of mathematical statistics}, 33\penalty0 (3):\penalty0 1065--1076, 1962.

\bibitem[Reiss and Hoshen(2023)]{reiss2023mean}
T.~Reiss and Y.~Hoshen.
\newblock Mean-shifted contrastive loss for anomaly detection.
\newblock In \emph{Proceedings of the AAAI Conference on Artificial Intelligence}, volume~37, pages 2155--2162, 2023.

\bibitem[Robinson et~al.(2020)Robinson, Chuang, Sra, and Jegelka]{robinson2020contrastive}
J.~D. Robinson, C.-Y. Chuang, S.~Sra, and S.~Jegelka.
\newblock Contrastive learning with hard negative samples.
\newblock In \emph{International Conference on Learning Representations}, 2020.

\bibitem[Roth et~al.(2022)Roth, Pemula, Zepeda, Sch{\"o}lkopf, Brox, and Gehler]{roth2022towards}
K.~Roth, L.~Pemula, J.~Zepeda, B.~Sch{\"o}lkopf, T.~Brox, and P.~Gehler.
\newblock Towards total recall in industrial anomaly detection.
\newblock In \emph{Proceedings of the IEEE/CVF conference on computer vision and pattern recognition}, pages 14318--14328, 2022.

\bibitem[Ruff et~al.(2018)Ruff, Vandermeulen, Goernitz, Deecke, Siddiqui, Binder, M{\"u}ller, and Kloft]{ruff2018deep}
L.~Ruff, R.~Vandermeulen, N.~Goernitz, L.~Deecke, S.~A. Siddiqui, A.~Binder, E.~M{\"u}ller, and M.~Kloft.
\newblock Deep one-class classification.
\newblock In \emph{Proceedings of the 35th International Conference on Machine Learning}, volume~80 of \emph{Proceedings of Machine Learning Research}, pages 4393--4402. PMLR, 10--15 Jul 2018.

\bibitem[Ruff et~al.(2020)Ruff, Vandermeulen, G{\"o}rnitz, Binder, M{\"u}ller, M{\"u}ller, and Kloft]{ruff2019deep}
L.~Ruff, R.~A. Vandermeulen, N.~G{\"o}rnitz, A.~Binder, E.~M{\"u}ller, K.-R. M{\"u}ller, and M.~Kloft.
\newblock Deep semi-supervised anomaly detection.
\newblock In \emph{International Conference on Learning Representations}, 2020.

\bibitem[Schl{\"u}ter et~al.(2022)Schl{\"u}ter, Tan, Hou, and Kainz]{schluter2022natural}
H.~M. Schl{\"u}ter, J.~Tan, B.~Hou, and B.~Kainz.
\newblock Natural synthetic anomalies for self-supervised anomaly detection and localization.
\newblock In \emph{European Conference on Computer Vision}, pages 474--489. Springer, 2022.

\bibitem[Sehwag et~al.(2021)Sehwag, Chiang, and Mittal]{sehwag2021ssd}
V.~Sehwag, M.~Chiang, and P.~Mittal.
\newblock Ssd: A unified framework for self-supervised outlier detection.
\newblock In \emph{International Conference on Learning Representations}, 2021.

\bibitem[Sohn(2016)]{sohn2016improved}
K.~Sohn.
\newblock Improved deep metric learning with multi-class n-pair loss objective.
\newblock \emph{Advances in neural information processing systems}, 29, 2016.

\bibitem[Sohn et~al.(2021)Sohn, Li, Yoon, Jin, and Pfister]{learn_and_eval}
K.~Sohn, C.-L. Li, J.~Yoon, M.~Jin, and T.~Pfister.
\newblock Learning and evaluating representations for deep one-class classification.
\newblock In \emph{International Conference on Learning Representations}, 2021.

\bibitem[Sun et~al.(2022)Sun, Ming, Zhu, and Li]{sun2022out}
Y.~Sun, Y.~Ming, X.~Zhu, and Y.~Li.
\newblock Out-of-distribution detection with deep nearest neighbors.
\newblock In \emph{International Conference on Machine Learning}, pages 20827--20840. PMLR, 2022.

\bibitem[Tack et~al.(2020)Tack, Mo, Jeong, and Shin]{csi}
J.~Tack, S.~Mo, J.~Jeong, and J.~Shin.
\newblock Csi: Novelty detection via contrastive learning on distributionally shifted instances.
\newblock In \emph{Advances in Neural Information Processing Systems}, volume~33, pages 11839--11852. Curran Associates, Inc., 2020.

\bibitem[Tian et~al.(2024)Tian, Fan, Isola, Chang, and Krishnan]{supcon2}
Y.~Tian, L.~Fan, P.~Isola, H.~Chang, and D.~Krishnan.
\newblock Stablerep: Synthetic images from text-to-image models make strong visual representation learners.
\newblock \emph{Advances in Neural Information Processing Systems}, 36, 2024.

\bibitem[Wang et~al.(2023{\natexlab{a}})Wang, Wang, Qin, Zhang, Bao, and Huang]{wang2023uniconha}
G.~Wang, Y.~Wang, J.~Qin, D.~Zhang, X.~Bao, and D.~Huang.
\newblock Unilaterally aggregated contrastive learning with hierarchical augmentation for anomaly detection.
\newblock In \emph{Proceedings of the IEEE/CVF International Conference on Computer Vision (ICCV)}, pages 6888--6897, October 2023{\natexlab{a}}.

\bibitem[Wang et~al.(2023{\natexlab{b}})Wang, Fang, Zhang, Liu, Li, and Han]{wang2023learning}
Q.~Wang, Z.~Fang, Y.~Zhang, F.~Liu, Y.~Li, and B.~Han.
\newblock Learning to augment distributions for out-of-distribution detection.
\newblock \emph{Advances in neural information processing systems}, 36:\penalty0 73274--73286, 2023{\natexlab{b}}.

\bibitem[Wu et~al.(2024)Wu, Mo, Feng, Atito, Kitler, and Awais]{wu2024rethinking}
J.~Wu, S.~Mo, Z.~Feng, S.~Atito, J.~Kitler, and M.~Awais.
\newblock Rethinking positive pairs in contrastive learning.
\newblock \emph{arXiv preprint arXiv:2410.18200}, 2024.

\bibitem[Wu et~al.(2018)Wu, Xiong, Yu, and Lin]{wu2018unsupervised}
Z.~Wu, Y.~Xiong, S.~X. Yu, and D.~Lin.
\newblock Unsupervised feature learning via non-parametric instance discrimination.
\newblock In \emph{Proceedings of the IEEE conference on computer vision and pattern recognition}, pages 3733--3742, 2018.

\bibitem[Xiao et~al.(2017)Xiao, Rasul, and Vollgraf]{fmnist_dataset}
H.~Xiao, K.~Rasul, and R.~Vollgraf.
\newblock Fashion-mnist: a novel image dataset for benchmarking machine learning algorithms.
\newblock \emph{CoRR}, abs/1708.07747, 2017.

\bibitem[Yang et~al.(2024)Yang, Zhou, Li, and Liu]{yang2024generalized}
J.~Yang, K.~Zhou, Y.~Li, and Z.~Liu.
\newblock Generalized out-of-distribution detection: A survey.
\newblock \emph{International Journal of Computer Vision}, pages 1--28, 2024.

\bibitem[Yi and Yoon(2021)]{yi2020psvdd}
J.~Yi and S.~Yoon.
\newblock Patch svdd: Patch-level svdd for anomaly detection and segmentation.
\newblock In H.~Ishikawa, C.-L. Liu, T.~Pajdla, and J.~Shi, editors, \emph{Computer Vision -- ACCV 2020}, pages 375--390, Cham, 2021. Springer International Publishing.

\bibitem[Zavrtanik et~al.(2021)Zavrtanik, Kristan, and Sko{\v{c}}aj]{zavrtanik2021draem}
V.~Zavrtanik, M.~Kristan, and D.~Sko{\v{c}}aj.
\newblock Draem-a discriminatively trained reconstruction embedding for surface anomaly detection.
\newblock In \emph{Proceedings of the IEEE/CVF international conference on computer vision}, pages 8330--8339, 2021.

\bibitem[Zhang et~al.(2023{\natexlab{a}})Zhang, Li, Li, Dai, Jiang, and Xia]{zhang2023unsupervised}
X.~Zhang, N.~Li, J.~Li, T.~Dai, Y.~Jiang, and S.-T. Xia.
\newblock Unsupervised surface anomaly detection with diffusion probabilistic model.
\newblock In \emph{Proceedings of the IEEE/CVF International Conference on Computer Vision}, pages 6782--6791, 2023{\natexlab{a}}.

\bibitem[Zhang et~al.(2023{\natexlab{b}})Zhang, Li, Li, Huang, Shan, and Chen]{zhang2023destseg}
X.~Zhang, S.~Li, X.~Li, P.~Huang, J.~Shan, and T.~Chen.
\newblock Destseg: Segmentation guided denoising student-teacher for anomaly detection.
\newblock In \emph{Proceedings of the IEEE/CVF Conference on Computer Vision and Pattern Recognition}, pages 3914--3923, 2023{\natexlab{b}}.

\bibitem[Zhang et~al.(2024)Zhang, Xu, and Zhou]{zhang2024realnet}
X.~Zhang, M.~Xu, and X.~Zhou.
\newblock Realnet: A feature selection network with realistic synthetic anomaly for anomaly detection.
\newblock In \emph{Proceedings of the IEEE/CVF conference on computer vision and pattern recognition}, pages 16699--16708, 2024.

\end{thebibliography}

\newpage
\appendix
\onecolumn
\section{Training Details}\label{app:training_details}

We use ResNet-18~\citep{he2016deep} as the encoder network \( f_\theta \), followed by a multi-layer perceptron projection head \( g_\phi \). The projection head consists of 9 layers: 8 hidden layers with 512 units each, followed by an output layer with 128 units. Formally, it is structured as:
\[
\underbrace{[512, 512, \dots, 512]}_{\text{8 layers}}, 128
\]
with each hidden layer followed by batch normalization and ReLU activations. All models are trained by minimizing the FIRM loss with a temperature parameter of \( \tau = 0.2 \). We optimize the models using stochastic gradient descent (SGD) with a momentum of 0.9, training for 2000 epochs. The learning rate is linearly warmed up over the first 20 epochs to 0.01 and then follows a cosine annealing schedule. We apply L2 weight regularization with a coefficient of 0.0003. We use a batch size of 32 for semantic anomaly detection and a batch size of 16 for industrial anomaly detection. We employ standard augmentations~\citep{simclr} for semantic anomaly detection during training, including random resizing and cropping, color jitter, and Gaussian blur. 
For industrial anomaly detection, we use translation and color jitter for ablation experiments, while for the main results, no augmentation is applied and pairs are created using the patch strategy. Our code can be found in \url{https://github.com/willtl/firm}.

\section{Additional Score Functions}\label{app:score_functions}
While parametric methods such as Mahalanobis distance have been widely adopted for \ac{OOD} detection due to their strong empirical performance and simplicity in modeling class-conditional distributions, we focus on non-parametric scoring functions that more directly leverage the structure of learned embeddings without relying on distributional assumptions.

\subsection{Detection Score}
Following the cosine similarity detection score in Equation~\eqref{eq:score}, we denote $s_{\textup{con}}^*$ as the cosine similarity score scaled by the representation norm:
\begin{equation}\label{eq:score_norm}
s_{\textup{con}}^*(x, \{x_m\}, k) = s_{\textup{con}}(x, \{x_m\}, k) \cdot \left\Vert f_{\theta}(x)\right\Vert.
\end{equation}

\subsection{Ensemble Score}
While employing rotation transformations to generate synthetic outliers during training, we can enhance the detection score by incorporating these transformations during inference. Specifically, the shifting transformation score averages the cosine similarity across multiple rotated input versions. Formally, the score is defined as:
\begin{equation}\label{eq:shift_score}
    s_{\textup{shift}}(x, \{x_m\}, k) = \frac{1}{4} \sum_{\gamma \in \{0, 90, 180, 270\}} s_{\textup{con}}(R_\gamma(x), R_\gamma(\{x_m\}), k),
\end{equation}
where $R_\gamma(x)$ denotes the rotation of sample $x$ by $\gamma$ degrees, and $R_\gamma(\{x_m\})$ denotes the set $\{x_m\}$ with each element rotated by $\gamma$ degrees. Note that this scoring function is ineffective when synthetic outliers are sourced through \ac{OE}, since test-time rotation does not align meaningfully with such externally sampled outliers, even if rotation transformations are used during training.
 
To further improve robustness, we employ an ensemble score, $s_{\textup{ens}}$, which incorporates multiple random crops in addition to the shifting transformation score. For each rotation degree $\theta$, we perform ten random crops of $x$, scaling within $[0.5, 1]$, and average the scores. The ensemble score is given by:
\begin{equation}\label{eq:ens_score}
s_{\textup{ens}}(x, \{x_m\}, k) = \frac{1}{4} \sum_{\theta \in \{0, 90, 180, 270\}} \frac{1}{10} \sum_{i=1}^{10} s_{\textup{con}}(C_i(R_\theta(x)), R_\theta(\{x_m\}), k),
\end{equation}
where $C_i$ denotes the $i$-th random crop applied after the rotation $R_\theta$. This ensemble strategy enhances detection by considering both rotational and spatial variations in the input.

\subsection{Prototype Score}
We introduce a prototype-based scoring function, which measures the cosine similarity between the sample and the prototype of the learned \ac{ID} representations, defined as 
\begin{equation}
 s_{\textup{proto}}(x) = 1 - \cos(\phi(x), c), 
\end{equation}
where \( c \in \mathbb{R}^d \) is the prototype of the \ac{ID} representations, and $\cos$ denotes the cosine similarity between $\phi(x)$ and $c$.

\subsection{Kernel Density Estimation}\label{app:oneclass_classifiers}

For completeness, we provide the formulation of the classical method \ac{KDE}~\citep{parzen1962estimation} employed in our ablation studies, which is commonly used for anomaly detection and related tasks. 
KDE is a non-parametric technique to estimate the probability density function of the input data distribution.
Following the notation from~\citet{learn_and_eval}, the normality score using KDE with a Radial Basis Function (RBF) kernel, parameterized by $\gamma$, is formulated as follows:
\begin{equation}
    \text{KDE}_{\gamma}(x) = -\frac{1}{\gamma} \log \left[ \sum_{y} \exp \left( -\gamma \|x - y\|^2 \right) \right].
\end{equation}  

Classical methods like \ac{KDE}~\citep{parzen1962estimation} are often used for anomaly detection and related tasks. However, these methods face significant challenges when dealing with high-dimensional data spaces due to the curse of dimensionality~\citep{ghosal2024overcome}. This issue affects their learning efficiency and generalization capabilities, making it difficult for these models to differentiate effectively between \ac{ID} and \ac{OOD} samples. In both discriminative and generative approaches, the primary challenge lies in learning compact latent representations that effectively capture the data manifold of \ac{ID} samples~\citep{learn_and_eval}.  

\section{Additional Ablation Studies}\label{app:ablation_studies}

\subsection{Encouraging Separation for Synthetic Outliers}\label{app:sucon_comparison}

For this experiment, we consider dataset \(\mathcal{D} = \mathcal{D}_{\text{in}} \cup \mathcal{D}_{\textup{sout}} = \{x_i\}_{i=1}^N \cup \{x_j\}_{j=1}^M\)\footnote{\(\mathcal{D}_{\text{in}}\) consists of \ac{ID} samples; \(\mathcal{D}_{\textup{sout}}\) consists of synthetic outliers.}, where \(x_j \in \mathcal{D}_{\textup{sout}}\) is generated by applying geometric transformations to \ac{ID} samples—specifically, rotations of \(90^\circ\), \(180^\circ\), and \(270^\circ\). In our experiment, we compare two strategies for defining the positive set \(\Pos(i)\) for synthetic outliers. In the first strategy, we treat synthetic outliers as a single class, defining the label space as \(\mathcal{Y} = \{1, 2\}\), where \(y = 1\) corresponds to normal samples (\ac{ID} samples), and \(y = 2\) denotes all synthetic outliers. In this strategy, the positive set for any synthetic outlier \(j\) is simply all synthetic outliers in the batch, i.e.,
   \[
   \Pos(j) = \{ k \in \mathcal{B} \mid y_k = 2 \} \setminus \{j\},
   \]
treating all synthetic outliers as interchangeable positives.  In the second strategy, we refine the label space to \(\mathcal{Y} = \{1, 2, 3, 4\}\), where each synthetic outlier group corresponding to a specific geometric transformation (rotation of \(90^\circ, 180^\circ, 270^\circ\)) is assigned a unique label. Specifically, for a synthetic outlier \(j\) with transformation \(T_j\), the positive set is restricted to synthetic outliers that share the same transformation type, i.e.,
\[
\Pos(j) = \{ k \in \mathcal{B} \mid y_k = y_j \} \setminus \{j\},
\]
where \(y_j\) corresponds to the label associated with the geometric transformation \(T_j\), and \(y_j > 1\) indicates the sample is a synthetic outlier, while \(y_j = 1\) denotes a normal sample. The experiment is conducted on CIFAR-10, where each class in the dataset is treated as normal. We report the anomaly detection performance of both strategies using three scoring functions: (1) the standard contrastive score \(\scon\), (2) the prototype-based scoring function \(s_{\textup{proto}}\), and (3) Kernel Density Estimation (KDE).  We observe that, in the single-class positive set strategy (Strategy 1), SupCon tends to encourage compactness for both \ac{ID} samples and synthetic outliers, which can result in representational collapse when synthetic outliers are treated as interchangeable positives. Specifically, we see that this approach performs suboptimally for well-known challenging classes such as the ``Bird'' and ``Cat'' categories, achieving scores of 65.8\% and 63.7\%, respectively, as shown in Table~\ref{tab:supcon_comparison}. In contrast, when applying the transformation-aware positive set strategy (Strategy 2), the model achieves significantly better results, with scores of 90.3\% and 80.3\% for the same classes. 

\begin{table}[!ht]
        \centering
        \scriptsize
        \setlength{\tabcolsep}{0pt}
        \caption{AUROC (\%) comparison of different scoring functions on SupCon and Rot-SupCon for CIFAR-10.}\label{tab:supcon_comparison}
        \begin{tabularx}{\textwidth}{@{} >{\hsize=2cm}L >{\hsize=1.5cm}X LLLLLLLLLLC@{}}
            \toprule
            \textbf{Loss} & \textbf{Score} & \textbf{Plane} & \textbf{Car} & \textbf{Bird} & \textbf{Cat} & \textbf{Deer} & \textbf{Dog} & \textbf{Frog} & \textbf{Horse} & \textbf{Ship} & \textbf{Truck} & \textbf{Mean} \\ 
            \midrule 
            SupCon & $\scon$ & 84.8\sd{0.0} & 97.8\sd{0.1} & 65.8\sd{0.5} & 63.7\sd{0.2} & 90.5\sd{0.2} & 88.5\sd{0.3} & 91.9\sd{0.1} & 97.1\sd{0.0} & 95.1\sd{0.1} & 94.7\sd{0.0} & 87.0\sd{0.1} \\
            SupCon & $s_{\textup{proto}}$ & 83.0\sd{0.4} & 97.1\sd{0.1} & 61.4\sd{0.6} & 59.4\sd{0.1} & 89.0\sd{0.9} & 88.2\sd{0.1} & 90.4\sd{0.1} & 96.4\sd{0.0} & 94.4\sd{0.3} & 94.1\sd{0.1} & 85.3\sd{0.3} \\
            SupCon & $\textup{KDE}$ & 83.5\sd{0.0} & 97.4\sd{0.1} & 63.8\sd{0.2} & 59.7\sd{0.2} & 90.2\sd{0.4} & 88.9\sd{0.1} & 91.2\sd{0.1} & 96.8\sd{0.1} & 94.9\sd{0.1} & 94.4\sd{0.1} & 86.1\sd{0.1} \\
            \midrule 
            Rot-SupCon & $\scon$ & 86.6\sd{0.0} & 98.2\sd{0.1} & 90.3\sd{0.1} & 80.3\sd{0.4} & 92.8\sd{0.3} & 92.2\sd{0.3} & 94.9\sd{0.0} & 98.1\sd{0.1} & 95.9\sd{0.1} & 95.5\sd{0.0} & 92.5\sd{0.1} \\
            Rot-SupCon & $s_{\textup{proto}}$ & 85.4\sd{0.1} & 97.7\sd{0.1} & 89.7\sd{0.2} & 80.1\sd{0.3} & 91.2\sd{0.5} & 91.9\sd{0.2} & 94.0\sd{0.1} & 97.4\sd{0.2} & 94.9\sd{0.2} & 94.8\sd{0.1} & 91.7\sd{0.2} \\
            Rot-SupCon & $\textup{KDE}$ & 85.4\sd{0.3} & 98.0\sd{0.1} & 89.9\sd{0.2} & 80.4\sd{0.1} & 92.1\sd{0.5} & 92.2\sd{0.2} & 94.5\sd{0.1} & 97.8\sd{0.1} & 95.4\sd{0.1} & 95.3\sd{0.0} & 92.1\sd{0.2} \\
            \bottomrule
        \end{tabularx}
\end{table} 
    
\subsection{Analysis of Score Functions and ID Compactness}\label{app:ablation_score_functions}
We assess the impact of different scoring functions on NT-Xent, SupCon, Rot-SupCon, and FIRM. The scoring functions include those from Section~\ref{sec:multi_positive} and Appendix~\ref{app:score_functions}. To evaluate the effectiveness of ensemble scores with \ac{OE}, we analyze ``FIRM + OE (rot.),'' where rotations are applied to synthetic outliers during training. Table~\ref{tab:ablation_scores} shows the results across scoring functions. FIRM improves notably with ensemble-based scores, though the inclusion of norms (e.g., $\scon^*$, $\sens^*$) does not significantly improve results. The prototype-based score performs well for FIRM and shows mixed results for Rot-SupCon.
Results with \ac{OE} suggest that applying rotations degrade the performance with respect to $\scon$, and offer limited or no gains for $s_{\textup{proto}}$.
``FIRM + OE'' achieves a competitive result with $s_{\textup{proto}}$, highlighting the importance of choosing effective scoring functions for anomaly detection.

\begin{table}[!h]
    \centering
    \scriptsize
    \setlength{\tabcolsep}{2pt}
    \caption{AUROC (\%) of different scoring functions applied to NT-Xent, SupCon, Rot-SupCon, and FIRM on CIFAR-10. We report absolute scores for $\scon$, and for all other scoring functions, we provide relative improvements. Each score variant (excluding $\scon$) is split into two subcolumns: \emph{$\Delta$Min} indicates percentage improvement over the minimum value in the respective column (i.e., relative performance within the scoring function), and \emph{$\Delta\scon$} shows percentage change relative to that method’s own $\scon$ score (i.e., performance change when switching from cosine to the given scoring method).}
    \label{tab:ablation_scores}
    \begin{tabularx}{\textwidth}{@{}>{\hsize=2.5cm}L C *{6}{CC}@{}}
        \toprule
        \textbf{Loss} & $\scon$ &
        \multicolumn{2}{c}{$\scon^*$} &
        \multicolumn{2}{c}{$\sshift$} &
        \multicolumn{2}{c}{$\sshift^*$} &
        \multicolumn{2}{c}{$\sens$} &
        \multicolumn{2}{c}{$\sens^*$} &
        \multicolumn{2}{c}{$s_{\textup{proto}}$} \\
        \cmidrule(lr){3-4}
        \cmidrule(lr){5-6}
        \cmidrule(lr){7-8}
        \cmidrule(lr){9-10}
        \cmidrule(lr){11-12}
        \cmidrule(lr){13-14}
        & & $\Delta$Min & $\Delta\scon$ & $\Delta$Min & $\Delta\scon$ & $\Delta$Min & $\Delta\scon$ & $\Delta$Min & $\Delta\scon$ & $\Delta$Min & $\Delta\scon$ & $\Delta$Min & $\Delta\scon$ \\
        \midrule
        NT-Xent & 92.2 & \showdiff{9.9} & \showdiff{-0.3} & \showdiff{6.1} & \showdiff{0.4} & \showdiff{8.7} & \showdiff{-0.7} & \showdiff{6.2} & \showdiff{0.8} & \showdiff{16.1} & \showdiff{0.0} & 84.2 & \showdiff{-8.7} \\
        SupCon & 86.5 & 83.6 & \showdiff{-3.4} & 87.3 & \showdiff{0.9} & 84.3 & \showdiff{-2.5} & 87.5 & \showdiff{1.2} & 79.4 & \showdiff{-8.2} & \showdiff{1.3} & \showdiff{-1.4} \\
        Rot-SupCon & 92.5 & \showdiff{10.4} & \showdiff{-0.2} & \showdiff{7.1} & \showdiff{1.1} & \showdiff{11.0} & \showdiff{1.2} & \showdiff{7.2} & \showdiff{1.4} & \showdiff{18.1} & \showdiff{1.4} & \showdiff{8.9} & \showdiff{-0.9} \\
        FIRM & 93.4 & \showdiff{11.7} & \showdiff{0.0} & \showdiff{8.8} & \showdiff{1.7} & \showdiff{12.7} & \showdiff{1.7} & \showdiff{8.9} & \showdiff{2.0} & \showdiff{20.2} & \showdiff{2.1} & \showdiff{10.2} & \showdiff{-0.6} \\
        \midrule
        FIRM + OE (rot.) & 97.0 & \showdiff{13.5} & \showdiff{-2.2} & \showdiff{9.6} & \showdiff{-1.3} & \showdiff{8.4} & \showdiff{-5.8} & \showdiff{9.1} & \showdiff{-1.5} & \showdiff{11.5} & \showdiff{-8.8} & \showdiff{15.4} & \showdiff{0.2} \\
        FIRM + OE & 97.4 & \showdiff{13.9} & \showdiff{-2.3} & \showdiff{10.1} & \showdiff{-1.3} & \showdiff{8.4} & \showdiff{-6.2} & \showdiff{9.6} & \showdiff{-1.5} & \showdiff{12.0} & \showdiff{-8.7} & \showdiff{15.8} & \showdiff{0.1} \\
        \bottomrule
    \end{tabularx}
\end{table}

The superior performance of FIRM with the prototype-based scoring function, $s_{\textup{proto}}$, as presented in Table~\ref{tab:ablation_scores}, yields two key outcomes. First, it highlights the importance of high-quality synthetic outliers—sourced in this instance from the 80 Million Tiny Images dataset~\citep{outlier_exposure}—that effectively approximate the true distribution of anomalies. Second, FIRM successfully leverages the diversity of these synthetic outliers to promote ID compactness. The effectiveness of FIRM in utilizing this information is reflected in its ability to use the prototype (center) of the ID representations, which is tightly linked to ID compactness, as a reference for computing distances and accurately identifying anomalies.

\subsection{Source of Synthetic Outliers}\label{app:source_of_synthetic_outliers}
We further investigate the impact of synthetic outliers on the performance of FIRM, NT-Xent, and SupCon. Table~\ref{tab:ablation_synthetic_outliers} presents results for CIFAR-10, comparing two settings: (a) synthetic outliers generated through rotations and (b) outliers from \ac{OE}. In the first setting, FIRM consistently outperforms NT-Xent and SupCon, particularly excelling in challenging classes like ``Bird'' and ``Cat.'' However, classes such as ``Plane'' and ``Car'' exhibit lower performance due to insufficient perceptual differences introduced by synthetic outliers derived from rotations. This limitation can lead to class collisions, where hard negatives overlap with ID samples. This issue is exacerbated by the multi-positive strategy. In contrast, when synthetic outliers are introduced via \ac{OE}, these performance issues, especially in classes like ``Plane'' and ``Car,'' are no longer present.

\begin{table}[!h]
    \centering
    \scriptsize
    \renewcommand{\arraystretch}{1.3}  
    \caption{Per-class AUROC (\%) comparison for CIFAR-10 using different sources of synthetic outliers: (a) synthetic outliers generated through rotations, and (b) synthetic outliers sourced via \ac{OE}. We use cosine similarity score (\( s_{\text{con}} \)) with \( k = 1 \). The relative improvement of FIRM over NT-Xent is shown in the bottom row.}
    \label{tab:ablation_synthetic_outliers}

    \textbf{(a)} Rotational Synthetic Outliers\\[0.3em]
    \begin{tabularx}{\textwidth}{@{} >{\hsize=2.4cm}L *{10}{C} C @{}}
            \toprule
            \textbf{Method} & \textbf{Plane} & \textbf{Car} & \textbf{Bird} & \textbf{Cat} & \textbf{Deer} & \textbf{Dog} & \textbf{Frog} & \textbf{Horse} & \textbf{Ship} & \textbf{Truck} & \textbf{Mean} \\
            \midrule
            NT-Xent & \textbf{90.8\sd{0.3}} & \textbf{98.7\sd{0.0}} & 89.0\sd{0.0} & 81.7\sd{0.3} & 91.0\sd{0.3} & 89.4\sd{0.3} & 93.4\sd{0.3} & 97.8\sd{0.0} & 95.4\sd{0.4} & 95.0\sd{0.5} & 92.2\sd{0.2} \\ 
            SupCon & 85.0\sd{0.1} & 97.7\sd{0.0} & 65.6\sd{0.2} & 62.0\sd{0.6} & 89.7\sd{0.0} & 87.8\sd{0.2} & 91.3\sd{0.2} & 96.8\sd{0.1} & 95.0\sd{0.1} & 94.3\sd{0.2} & 86.5\sd{0.2} \\
            Rot-SupCon & 87.0\sd{0.0} & 98.1\sd{0.0} & 90.4\sd{0.2} & 80.3\sd{0.3} & 92.8\sd{0.3} & 92.1\sd{0.3} & \textbf{94.9\sd{0.0}} & \textbf{98.1\sd{0.2}} & 96.0\sd{0.1} & \textbf{95.5\sd{0.1}} & 92.5\sd{0.2} \\
            $\FIRM$ & 89.7\sd{0.5} & 98.4\sd{0.0} & \textbf{91.4\sd{0.2}} & \textbf{83.9\sd{0.7}} & \textbf{93.6\sd{0.0}} & \textbf{92.7\sd{0.2}} & 94.5\sd{0.2} & \textbf{98.1\sd{0.0}} & \textbf{96.6\sd{0.1}} & 95.4\sd{0.0} & \textbf{93.4\sd{0.2}} \\
            \midrule
            Relative Improvement & \showdiff{-1.2} & \showdiff{-0.3} & \showdiff{2.7} & \showdiff{2.7} & \showdiff{2.9} & \showdiff{3.7} & \showdiff{1.2} & \showdiff{0.3} & \showdiff{1.3} & \showdiff{0.4} & \showdiff{1.4} \\ 
            \bottomrule
    \end{tabularx}

    \vspace{1.2em}  
 
    \textbf{(b)} Synthetic Outliers from \ac{OE}\\[0.3em]
    \begin{tabularx}{\textwidth}{@{} >{\hsize=2.4cm}L *{10}{C} C @{}}
            \toprule
            \textbf{Method} & \textbf{Plane} & \textbf{Car} & \textbf{Bird} & \textbf{Cat} & \textbf{Deer} & \textbf{Dog} & \textbf{Frog} & \textbf{Horse} & \textbf{Ship} & \textbf{Truck} & \textbf{Mean} \\
            \midrule 
            Rot. + OE & 90.4 & 99.3 & 93.7 & 88.1 & 97.4 & 94.3 & 97.1 & 98.8 & 98.7 & 98.5 & 95.6  \\
            \midrule
            NT-Xent + \ac{OE} & 91.6\sd{0.1} & 98.5\sd{0.0} & 87.5\sd{0.9} & 78.9\sd{0.2} & 90.3\sd{0.9} & 88.3\sd{0.1} & 95.8\sd{0.1} & 95.0\sd{0.1} & 96.0\sd{0.2} & 95.0\sd{0.2} & 91.7\sd{0.3} \\ 
            SupCon + OE & 96.7\sd{0.0} & 98.4\sd{0.1} & 91.0\sd{0.9} & 86.7\sd{0.9} & 97.5\sd{0.6} & 93.0\sd{0.2} & 97.8\sd{0.1} & 97.7\sd{0.1} & 98.0\sd{0.0} & 98.1\sd{0.1} & 95.5\sd{0.3} \\ 
            $\FIRM$ + \ac{OE} & \textbf{97.6\sd{0.2}} & \textbf{99.2\sd{0.0}} & \textbf{95.8\sd{0.0}} & \textbf{92.1\sd{0.3}} & \textbf{98.1\sd{0.0}} & \textbf{96.2\sd{0.0}} & \textbf{98.8\sd{0.0}} & \textbf{98.6\sd{0.1}} & \textbf{98.8\sd{0.0}} & \textbf{98.8\sd{0.0}} & \textbf{97.4\sd{0.1}} \\
            \midrule
            Relative Improvement & \showdiff{6.6} & \showdiff{0.7} & \showdiff{9.5} & \showdiff{16.7} & \showdiff{8.6} & \showdiff{8.9} & \showdiff{3.1} & \showdiff{3.8} & \showdiff{2.9} & \showdiff{4.0} & \showdiff{6.2} \\ 
            \bottomrule
    \end{tabularx}
\end{table}

These findings highlight the importance of selecting appropriate synthetic outliers for anomaly detection. For instance, in the case of plane images, applying rotations (e.g., \(90^\circ, 180^\circ\)) does not generate meaningful outliers that lie outside the \ac{ID} distribution, but rather creates variations within the same distribution (\(P_{\textup{in}}\)), which do not effectively approximate the low-density boundary of \ac{ID} samples. These rotations fail to provide the perceptual diversity needed to challenge the model, causing synthetic outliers to overlap with ID samples. This emphasizes that synthetic outliers must be carefully designed to approximate the boundary of \(P_{\textup{in}}\), as described in Section~\ref{sec:synthetic_outliers}, ensuring that they lie truly outside the ID distribution, thereby providing the necessary contrast for effective representation learning.

\subsection{Temperature and Batch Size Analysis}\label{app:temp_batch}
\begin{figure*}[!h]
    \scriptsize
    \centering
    \setlength{\tabcolsep}{0pt}
    \begin{tabular}{cccc} \includegraphics[width=0.25\textwidth]{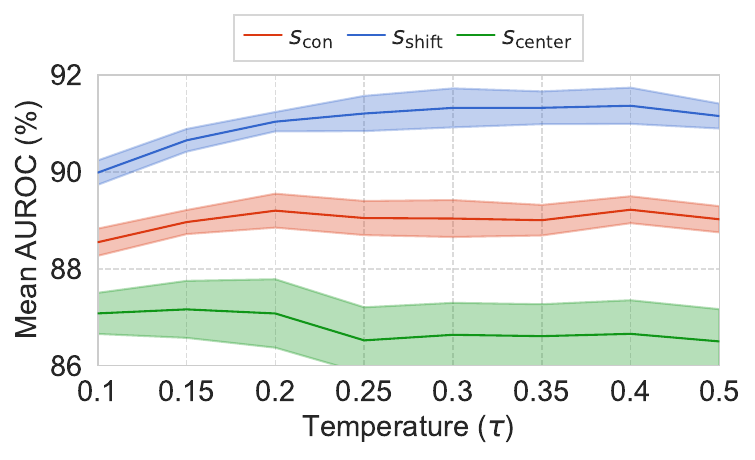} &  \includegraphics[width=0.25\textwidth]{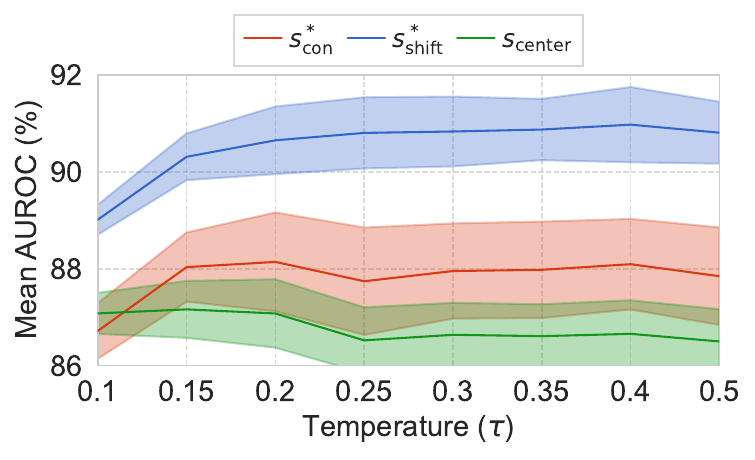} &  \includegraphics[width=0.25\textwidth]{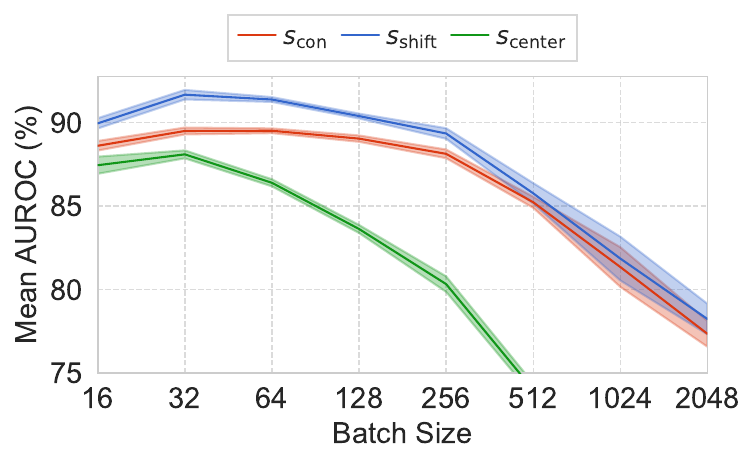} &  \includegraphics[width=0.25\textwidth]{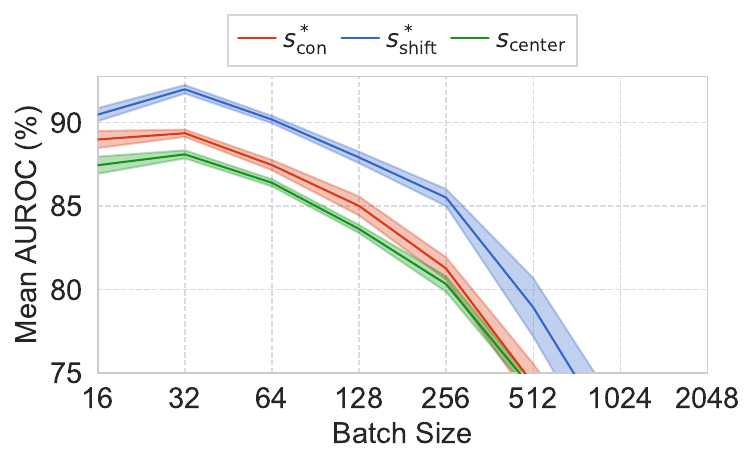}  \\
    (a) Temperature without norm. & (a) Temperature analysis with norm. & (c) Batch size without norm.  & (d) Bath size with norm.
    \end{tabular} 
        \caption{Analysis on the impact of varying values for $\tau$ and batch size on the performance of $\FIRM$.}\label{fig:ablation_learning_curves}
\end{figure*}

We assess the performance of \ac{FIRM} under different temperature values $\tau$ and batch sizes, focusing on four challenging CIFAR-10 classes: plane (0), bird (2), cat (3), and deer (5). Temperatures range from $\tau \in \{0.1, 0.15, 0.2, 0.25, 0.3, 0.35, 0.4, 0.5\}$, and batch sizes from $\{16, 32, 64, 128, 256, 512, 1024, 2048\}$. To understand the impact of these hyperparameters, we evaluate scoring functions with and without normalization, addressing differences observed in previous work. Figure~\ref{fig:ablation_learning_curves} (a) and (b) show the mean and variance in \ac{AUROC} across temperatures. Performance improves sharply from $\tau = 0.1$ and peaks at $\tau = 0.2$, consistent with~\citet{learn_and_eval}. However, when norms are used in the score functions, performance peaks at $\tau = 0.4$, aligning with~\citet{csi}. This suggests that using $\tau = 0.2$, as in our experiments, may explain lower performance when norms are included, indicating a higher temperature is needed for optimal results with norms. Figures~\ref{fig:ablation_learning_curves} (c) and (d) show performance across batch sizes, with better results at smaller sizes, consistent with~\citet{learn_and_eval}. Performance drops with larger batch sizes, particularly when norms are included.  

\subsection{Results for Different Scoring Functions and Synthetic Outlier Sources}
We present the results for NT-Xent, SupCon, Rot-SupCon, and FIRM using multiple scoring functions across CIFAR-10 classes with both generated synthetic outliers and \ac{OE}. The scoring functions evaluated include those proposed in Section~\ref{sec:learning_robust} and Appendix~\ref{app:score_functions}, including scoring function $s_{\textup{proto}}$. Scores marked with $*$ incorporate the norm, following Equation~(\ref{eq:score_norm}). Table~\ref{tab:distance_per_class_scores} provides the results for synthetic outliers generated as described in Section~\ref{sec:learning_robust}, while Table~\ref{tab:distance_per_class_scores_oe} presents the results with external data from~\citet{outlier_exposure}, referred to as \ac{OE}.

\begin{table}[!ht]
        \centering
        \scriptsize
        \setlength{\tabcolsep}{0pt}
        \caption{AUROC (\%) comparison for NT-Xent, SupCon, Rot-SupCon, and FIRM on CIFAR-10 with synthetic outliers generated through rotations. Values are reported with $k=5$.}\label{tab:distance_per_class_scores}
        \begin{tabularx}{\textwidth}{@{} >{\hsize=2cm}L >{\hsize=1.5cm}X LLLLLLLLLLC@{}}
            \toprule
            \textbf{Loss} & \textbf{Score} & \textbf{Plane} & \textbf{Car} & \textbf{Bird} & \textbf{Cat} & \textbf{Deer} & \textbf{Dog} & \textbf{Frog} & \textbf{Horse} & \textbf{Ship} & \textbf{Truck} & \textbf{Mean} \\ 
            \midrule 
            NT-Xent & $\scon$ & 91.7\sd{0.4} & 98.9\sd{0.0} & 89.8\sd{0.2} & 83.4\sd{0.1} & 91.7\sd{0.2} & 90.5\sd{0.2} & 94.6\sd{0.4} & 98.0\sd{0.0} & 95.7\sd{0.3} & 95.6\sd{0.2} & 93.0\sd{0.2} \\
            NT-Xent & $\scon^*$ & 90.3\sd{0.8} & 99.0\sd{0.1} & 88.7\sd{0.4} & 82.4\sd{0.9} & 90.5\sd{0.8} & 90.1\sd{0.5} & 92.0\sd{0.5} & 97.5\sd{0.1} & 95.4\sd{0.6} & 94.9\sd{0.3} & 92.1\sd{0.5} \\
            NT-Xent & $\sshift$ & 92.5\sd{0.3} & 99.1\sd{0.0} & 88.6\sd{0.9} & 82.8\sd{0.2} & 91.8\sd{0.2} & 90.3\sd{0.2} & 95.1\sd{0.8} & 98.1\sd{0.2} & 96.4\sd{0.1} & 95.9\sd{0.1} & 93.1\sd{0.3} \\
            NT-Xent & $\sshift^*$ & 91.2\sd{0.1} & 99.0\sd{0.0} & 88.5\sd{0.3} & 81.4\sd{0.1} & 89.4\sd{0.3} & 88.9\sd{0.2} & 92.1\sd{0.5} & 97.3\sd{0.3} & 95.9\sd{0.2} & 94.8\sd{0.5} & 91.8\sd{0.2} \\
            NT-Xent & $\sens$ & 92.9\sd{0.2} & 99.1\sd{0.0} & 89.2\sd{0.8} & 83.2\sd{0.2} & 92.3\sd{0.2} & 90.0\sd{0.1} & 95.2\sd{0.9} & 98.1\sd{0.2} & 96.3\sd{0.1} & 95.8\sd{0.1} & 93.2\sd{0.3} \\
            NT-Xent & $\sens^*$ & 91.5\sd{0.1} & 99.1\sd{0.0} & 89.3\sd{0.4} & 82.5\sd{0.3} & 90.2\sd{0.2} & 89.5\sd{0.0} & 92.9\sd{1.0} & 97.7\sd{0.3} & 96.2\sd{0.1} & 95.1\sd{0.5} & 92.4\sd{0.3} \\
            NT-Xent & $s_{\textup{proto}}$ & 72.4\sd{1.1} & 93.5\sd{0.2} & 68.8\sd{1.1} & 72.6\sd{0.9} & 88.6\sd{0.8} & 83.2\sd{1.0} & 90.3\sd{1.8} & 94.5\sd{1.2} & 85.2\sd{3.5} & 92.6\sd{0.1} & 84.2\sd{1.2} \\
            \midrule 
            SupCon & $\scon$ & 84.8\sd{0.0} & 97.8\sd{0.1} & 65.8\sd{0.5} & 63.7\sd{0.2} & 90.5\sd{0.2} & 88.5\sd{0.3} & 91.9\sd{0.1} & 97.1\sd{0.0} & 95.1\sd{0.1} & 94.7\sd{0.0} & 87.0\sd{0.1} \\
            SupCon & $\scon^*$ & 84.5\sd{0.4} & 98.2\sd{0.0} & 59.8\sd{0.6} & 60.8\sd{0.4} & 82.7\sd{1.6} & 82.9\sd{0.4} & 84.4\sd{2.1} & 95.9\sd{0.3} & 94.5\sd{0.0} & 94.0\sd{0.0} & 83.8\sd{0.6} \\
            SupCon & $\sshift$ & 86.2\sd{0.1} & 97.6\sd{0.0} & 68.7\sd{0.7} & 62.7\sd{1.1} & 91.3\sd{0.2} & 88.7\sd{0.1} & 92.7\sd{0.2} & 96.9\sd{0.3} & 95.4\sd{0.0} & 95.0\sd{0.0} & 87.5\sd{0.3} \\
            SupCon & $\sshift^*$ & 86.6\sd{0.5} & 98.3\sd{0.1} & 59.3\sd{0.3} & 57.9\sd{0.3} & 84.3\sd{1.4} & 84.7\sd{0.8} & 89.8\sd{1.1} & 95.2\sd{0.5} & 93.9\sd{0.5} & 94.4\sd{0.1} & 84.4\sd{0.6} \\
            SupCon & $\sens$ & 86.3\sd{0.2} & 97.5\sd{0.0} & 68.3\sd{0.2} & 63.7\sd{0.1} & 91.1\sd{0.1} & 88.1\sd{0.1} & 93.6\sd{0.0} & 96.3\sd{0.0} & 95.3\sd{0.0} & 94.8\sd{0.0} & 87.5\sd{0.1} \\
            SupCon & $\sens^*$ & 77.9\sd{5.5} & 98.4\sd{0.0} & 46.0\sd{5.5} & 49.3\sd{4.8} & 81.3\sd{0.9} & 83.3\sd{2.3} & 78.1\sd{0.0} & 96.0\sd{0.0} & 91.2\sd{0.0} & 94.3\sd{0.0} & 79.6\sd{1.9} \\
            SupCon & $s_{\textup{proto}}$ & 83.0\sd{0.4} & 97.1\sd{0.1} & 61.4\sd{0.6} & 59.4\sd{0.1} & 89.0\sd{0.9} & 88.2\sd{0.1} & 90.4\sd{0.1} & 96.4\sd{0.0} & 94.4\sd{0.3} & 94.1\sd{0.1} & 85.3\sd{0.3} \\
            \midrule 
            Rot-SupCon & $\scon$ & 86.6\sd{0.0} & 98.2\sd{0.1} & 90.3\sd{0.1} & 80.3\sd{0.4} & 92.8\sd{0.3} & 92.2\sd{0.3} & 94.9\sd{0.0} & 98.1\sd{0.1} & 95.9\sd{0.1} & 95.5\sd{0.0} & 92.5\sd{0.1} \\
            Rot-SupCon & $\scon^*$ & 86.6\sd{0.3} & 98.4\sd{0.0} & 89.7\sd{0.3} & 81.8\sd{0.1} & 91.7\sd{0.1} & 91.6\sd{0.2} & 94.3\sd{0.0} & 98.0\sd{0.2} & 96.2\sd{0.0} & 94.9\sd{0.0} & 92.3\sd{0.1} \\
            Rot-SupCon & $\sshift$ & 87.0\sd{0.2} & 98.4\sd{0.0} & 92.2\sd{0.1} & 82.0\sd{0.0} & 93.8\sd{0.1} & 93.0\sd{0.2} & 96.4\sd{0.2} & 98.4\sd{0.0} & 96.6\sd{0.1} & 95.9\sd{0.1} & 93.4\sd{0.1} \\
            Rot-SupCon & $\sshift^*$ & 88.6\sd{0.1} & 98.7\sd{0.1} & 91.8\sd{0.3} & 84.3\sd{0.1} & 92.8\sd{0.3} & 92.7\sd{0.4} & 96.2\sd{0.0} & 98.5\sd{0.0} & 97.1\sd{0.0} & 95.4\sd{0.1} & 93.6\sd{0.1} \\
            Rot-SupCon & $\sens$ & 87.1\sd{0.6} & 98.4\sd{0.1} & 92.7\sd{0.2} & 82.7\sd{0.0} & 94.3\sd{0.0} & 93.0\sd{0.4} & 96.9\sd{0.0} & 98.5\sd{0.0} & 96.9\sd{0.0} & 95.9\sd{0.0} & 93.6\sd{0.1} \\
            Rot-SupCon & $\sens^*$ & 88.2\sd{0.5} & 98.6\sd{0.1} & 92.7\sd{0.1} & 84.4\sd{0.2} & 93.1\sd{0.1} & 92.9\sd{0.5} & 96.6\sd{0.0} & 98.5\sd{0.0} & 97.5\sd{0.0} & 95.2\sd{0.0} & 93.8\sd{0.1} \\
            Rot-SupCon & $s_{\textup{proto}}$ & 85.4\sd{0.1} & 97.7\sd{0.1} & 89.7\sd{0.2} & 80.1\sd{0.3} & 91.2\sd{0.5} & 91.9\sd{0.2} & 94.0\sd{0.1} & 97.4\sd{0.2} & 94.9\sd{0.2} & 94.8\sd{0.1} & 91.7\sd{0.2} \\
            \midrule 
            FIRM & $\scon$ & 89.2\sd{0.5} & 98.3\sd{0.0} & 91.6\sd{0.0} & 84.0\sd{0.7} & 93.7\sd{0.0} & 92.8\sd{0.3} & 94.8\sd{0.3} & 98.1\sd{0.0} & 96.6\sd{0.1} & 95.3\sd{0.0} & 93.4\sd{0.2} \\
            FIRM & $\scon^*$ & 89.0\sd{0.2} & 98.5\sd{0.1} & 91.4\sd{0.2} & 85.4\sd{0.1} & 92.9\sd{0.1} & 93.2\sd{0.2} & 94.7\sd{0.4} & 97.9\sd{0.0} & 96.6\sd{0.0} & 95.0\sd{0.2} & 93.5\sd{0.2} \\
            FIRM & $\sshift$ & 91.9\sd{0.0} & 99.1\sd{0.0} & 93.3\sd{0.1} & 87.0\sd{0.2} & 95.0\sd{0.0} & 94.0\sd{0.0} & 97.0\sd{0.1} & 98.8\sd{0.0} & 97.7\sd{0.0} & 96.7\sd{0.0} & 95.1\sd{0.0} \\
            FIRM & $\sshift^*$ & 92.4\sd{0.2} & 99.2\sd{0.0} & 93.2\sd{0.1} & 87.9\sd{0.1} & 94.1\sd{0.1} & 94.0\sd{0.2} & 96.3\sd{0.4} & 98.7\sd{0.1} & 97.9\sd{0.0} & 96.3\sd{0.0} & 95.0\sd{0.1} \\
            FIRM & $\sens$ & 92.6\sd{0.0} & 99.2\sd{0.0} & 93.9\sd{0.1} & 87.6\sd{0.1} & 95.4\sd{0.0} & 94.2\sd{0.0} & 97.4\sd{0.1} & 98.8\sd{0.0} & 97.7\sd{0.0} & 96.5\sd{0.0} & 95.3\sd{0.0} \\
            FIRM & $\sens^*$ & 93.3\sd{0.3} & 99.2\sd{0.0} & 93.5\sd{0.3} & 89.0\sd{0.1} & 94.6\sd{0.0} & 94.4\sd{0.2} & 96.9\sd{0.3} & 98.8\sd{0.0} & 98.1\sd{0.0} & 96.4\sd{0.0} & 95.4\sd{0.1} \\
            FIRM & $s_{\textup{proto}}$ & 86.5\sd{0.0} & 98.1\sd{0.0} & 90.2\sd{0.0} & 83.5\sd{0.6} & 93.1\sd{0.2} & 92.7\sd{0.4} & 95.0\sd{0.3} & 97.9\sd{0.0} & 96.0\sd{0.2} & 94.8\sd{0.1} & 92.8\sd{0.2} \\
            \bottomrule
        \end{tabularx}
\end{table}

In the experiments shown in Table~\ref{tab:distance_per_class_scores}, incorporating norms into the scoring functions within NT-Xent and SupCon configurations leads to consistently lower performance in $\scon$, $\sshift$, and $\sens$ methods. This pattern is not observed with Rot-SupCon and FIRM, where the effect of norms on performance varies. This suggests that while norms may constrain the feature space in binary scenarios like NT-Xent and SupCon, reducing their effectiveness. They can promote generalization and robustness in more complex or multiclass settings such as Rot-SupCon and FIRM. Within NT-Xent and SupCon, $\sens$ emerges as the most effective scoring function, while $\sens^*$ achieves the highest AUROC scores for Rot-SupCon and FIRM. Alternatively, the $s_{\textup{proto}}$ scoring function, described in Section~\ref{sec:ablation}, consistently shows the poorest performance across NT-Xent, Rot-SupCon, and FIRM.

\begin{table}[!ht]
        \centering
        \scriptsize
        \setlength{\tabcolsep}{0pt}
        \caption{AUROC (\%) comparison for NT-Xent, SupCon, and FIRM losses on CIFAR-10 with Outlier Exposure (OE).}\label{tab:distance_per_class_scores_oe}
        \begin{tabularx}{\textwidth}{@{} >{\hsize=2cm}L >{\hsize=1.5cm}X LLLLLLLLLLC@{}}
            \toprule
            \textbf{Loss} & \textbf{Score} & \textbf{Plane} & \textbf{Car} & \textbf{Bird} & \textbf{Cat} & \textbf{Deer} & \textbf{Dog} & \textbf{Frog} & \textbf{Horse} & \textbf{Ship} & \textbf{Truck} & Mean \\ 
            \midrule 
            NT-Xent + OE & $\scon$ & 92.5\sd{0.1} & 98.7\sd{0.0} & 88.0\sd{0.8} & 80.5\sd{0.4} & 91.4\sd{0.6} & 90.2\sd{0.4} & 96.5\sd{0.0} & 96.2\sd{0.0} & 96.2\sd{0.1} & 95.7\sd{0.2} & 92.6\sd{0.3} \\
            NT-Xent + OE & $\scon^*$ & 88.3\sd{0.9} & 98.0\sd{0.0} & 88.1\sd{0.5} & 75.0\sd{0.5} & 90.5\sd{0.1} & 85.0\sd{0.1} & 95.1\sd{0.2} & 94.4\sd{0.1} & 94.8\sd{0.1} & 92.9\sd{0.0} & 90.2\sd{0.2} \\
            NT-Xent + OE & $\sshift$ & 90.7\sd{0.1} & 97.5\sd{0.1} & 84.8\sd{0.1} & 78.3\sd{0.9} & 88.7\sd{0.1} & 87.9\sd{0.1} & 94.8\sd{0.1} & 94.3\sd{0.1} & 93.2\sd{0.2} & 93.7\sd{0.1} & 90.4\sd{0.2} \\
            NT-Xent + OE & $\sshift^*$ & 82.1\sd{1.5} & 94.8\sd{0.0} & 82.6\sd{0.1} & 69.4\sd{1.8} & 84.2\sd{1.7} & 78.8\sd{0.2} & 91.8\sd{0.7} & 90.4\sd{0.5} & 85.8\sd{0.3} & 84.4\sd{0.3} & 84.4\sd{0.7} \\
            NT-Xent + OE & $\sens$ & 89.9\sd{0.0} & 97.2\sd{0.0} & 84.2\sd{0.0} & 77.7\sd{0.9} & 88.5\sd{0.1} & 86.9\sd{0.5} & 94.5\sd{0.1} & 93.8\sd{0.3} & 93.0\sd{0.2} & 93.1\sd{0.1} & 89.9\sd{0.2} \\
            NT-Xent + OE & $\sens^*$ & 71.5\sd{0.8} & 92.7\sd{0.8} & 82.3\sd{0.2} & 70.4\sd{1.9} & 82.2\sd{3.8} & 74.1\sd{4.3} & 83.4\sd{3.6} & 88.0\sd{0.3} & 79.9\sd{5.8} & 82.1\sd{2.9} & 80.7\sd{2.4} \\
            NT-Xent + OE & $s_{\textup{proto}}$ & 88.2\sd{1.3} & 98.2\sd{0.1} & 75.6\sd{0.5} & 72.5\sd{0.7} & 88.8\sd{2.8} & 86.1\sd{0.2} & 96.7\sd{0.0} & 96.6\sd{0.6} & 96.1\sd{0.2} & 94.7\sd{0.0} & 89.4\sd{0.6} \\
            \midrule
            SupCon + OE & $\scon$ & 97.0\sd{0.0} & 98.8\sd{0.1} & 92.1\sd{1.1} & 88.6\sd{0.7} & 97.7\sd{0.5} & 93.7\sd{0.1} & 98.3\sd{0.0} & 98.1\sd{0.0} & 98.7\sd{0.1} & 98.6\sd{0.0} & 96.2\sd{0.3} \\
            SupCon + OE & $\scon^*$ & 93.2\sd{0.9} & 99.0\sd{0.1} & 84.1\sd{0.1} & 81.9\sd{1.2} & 93.8\sd{1.6} & 91.8\sd{1.1} & 96.3\sd{0.1} & 96.2\sd{0.5} & 95.2\sd{0.2} & 96.1\sd{0.1} & 92.8\sd{0.6} \\
            SupCon + OE & $\sshift$ & 96.1\sd{0.1} & 97.9\sd{0.5} & 89.3\sd{0.6} & 84.2\sd{0.9} & 96.5\sd{1.0} & 91.4\sd{0.3} & 97.3\sd{0.2} & 97.1\sd{0.3} & 98.1\sd{0.2} & 97.9\sd{0.0} & 94.6\sd{0.4} \\
            SupCon + OE & $\sshift^*$ & 88.8\sd{1.4} & 98.5\sd{0.2} & 74.9\sd{0.7} & 78.0\sd{2.9} & 88.4\sd{1.4} & 88.1\sd{1.5} & 93.8\sd{0.5} & 93.2\sd{0.0} & 88.4\sd{2.2} & 93.9\sd{0.7} & 88.6\sd{1.1} \\
            SupCon + OE & $\sens$ & 96.5\sd{0.1} & 97.7\sd{0.9} & 89.6\sd{0.8} & 83.9\sd{0.4} & 97.0\sd{0.8} & 90.6\sd{1.3} & 97.7\sd{0.1} & 97.2\sd{0.4} & 97.7\sd{0.6} & 97.8\sd{0.1} & 94.6\sd{0.6} \\
            SupCon + OE & $\sens^*$ & 83.8\sd{0.1} & 93.1\sd{5.3} & 68.4\sd{1.8} & 77.8\sd{1.3} & 69.9\sd{7.9} & 87.1\sd{0.3} & 88.1\sd{2.1} & 90.4\sd{0.2} & 58.2\sd{12.6} & 93.3\sd{1.5} & 81.0\sd{3.3} \\
            SupCon + OE & $s_{\textup{proto}}$ & 96.7\sd{0.1} & 98.9\sd{0.0} & 92.9\sd{0.5} & 89.8\sd{0.1} & 97.5\sd{0.4} & 94.3\sd{0.0} & 98.3\sd{0.0} & 98.2\sd{0.0} & 98.9\sd{0.1} & 98.6\sd{0.1} & 96.4\sd{0.1} \\
            \midrule
            FIRM + OE & $\scon$ & 97.7\sd{0.1} & 99.2\sd{0.0} & 96.1\sd{0.0} & 92.6\sd{0.1} & 98.2\sd{0.0} & 96.4\sd{0.1} & 98.9\sd{0.0} & 98.8\sd{0.0} & 98.9\sd{0.0} & 99.0\sd{0.0} & 97.6\sd{0.0} \\
            FIRM + OE & $\scon^*$ & 95.5\sd{0.4} & 98.9\sd{0.0} & 93.0\sd{0.7} & 87.5\sd{0.4} & 96.6\sd{0.0} & 94.3\sd{0.0} & 98.3\sd{0.2} & 96.5\sd{0.3} & 97.6\sd{0.2} & 96.8\sd{0.0} & 95.5\sd{0.2} \\
            FIRM + OE & $\sshift$ & 96.7\sd{0.0} & 98.7\sd{0.0} & 93.6\sd{0.3} & 88.7\sd{0.4} & 97.3\sd{0.1} & 94.7\sd{0.5} & 98.5\sd{0.1} & 97.8\sd{0.1} & 97.9\sd{0.0} & 98.2\sd{0.1} & 96.2\sd{0.2} \\
            FIRM + OE & $\sshift^*$ & 91.9\sd{0.4} & 98.2\sd{0.1} & 86.4\sd{1.3} & 81.9\sd{0.1} & 92.0\sd{0.0} & 89.9\sd{0.1} & 97.1\sd{0.2} & 91.4\sd{0.4} & 94.6\sd{0.2} & 93.2\sd{0.2} & 91.7\sd{0.3} \\
            FIRM + OE & $\sens$ & 96.6\sd{0.2} & 98.5\sd{0.0} & 93.6\sd{0.3} & 88.7\sd{0.7} & 97.3\sd{0.2} & 94.4\sd{0.5} & 98.5\sd{0.2} & 97.5\sd{0.1} & 97.6\sd{0.1} & 97.9\sd{0.1} & 96.1\sd{0.2} \\
            FIRM + OE & $\sens^*$ & 88.9\sd{3.3} & 93.9\sd{0.1} & 86.3\sd{1.8} & 83.7\sd{0.6} & 91.5\sd{0.2} & 88.5\sd{2.4} & 96.6\sd{1.0} & 90.9\sd{0.3} & 87.6\sd{7.0} & 87.5\sd{0.9} & 89.5\sd{1.8} \\
            FIRM + OE & $s_{\textup{proto}}$ & 97.4\sd{0.1} & 99.3\sd{0.0} & 95.9\sd{0.1} & 92.5\sd{0.2} & 97.9\sd{0.0} & 96.3\sd{0.0} & 99.0\sd{0.1} & 99.0\sd{0.0} & 99.0\sd{0.1} & 98.9\sd{0.0} & 97.5\sd{0.1} \\
            \bottomrule
        \end{tabularx}
\end{table}

In the \ac{OE} experiments detailed in Table~\ref{tab:distance_per_class_scores_oe}, including norms in the scoring functions $\scon$, $\sshift$, and $\sens$ consistently degrades performance for all three loss functions. In contrast, the $s_{\textup{proto}}$ scoring function, which previously showed the lowest performance in synthetic outlier scenarios for NT-Xent and FIRM, exhibits improved performance in OE, achieving the best performance for SupCon and the second-best for FIRM. Notably, the most effective scoring functions in OE scenarios are: $\scon$ for both NT-Xent and FIRM; and $s_{\textup{proto}}$ for SupCon, indicating a distinct interaction between the outlier data types and the scoring methods.

\section{Per-Class Results}\label{app:per_class_results}
To better understand the behavior of contrastive objectives across different semantic categories, we report detailed per-class AUROC results for all evaluated datasets. These results highlight class-wise variance in anomaly detection performance, which is particularly relevant for benchmarking methods that may perform unevenly across classes despite strong average performance.

We present results for the following datasets: CIFAR-10, CIFAR-100 (superclass setting), Fashion-MNIST, Cats-vs-Dogs, and MVTec-AD. Tables~\ref{tab:cipher10_per_class_classifiers}, \ref{tab:cipher100_per_class_classifiers}, \ref{tab:fmnist_per_class_classifiers}, \ref{tab:catsdogs_per_class_classifiers}, and \ref{tab:results_mvtec_perclass_full} report the full per-class AUROC scores under different scoring strategies ($\scon$, $\sshift$, $\sens$) and training objectives (NT-Xent, SupCon, Rot-SupCon, and FIRM). Results are reported with $k=5$ and include standard deviations. Note that, MVTec-AD results shown here correspond to the ablation setting described in Section~\ref{sec:ablation}, where FIRM is trained using NSA-generated synthetic outliers on full images, without employing the patch-based strategy or SIA. These results differ from the main MVTec-AD experiments reported in Section~\ref{sec:main_results}, which incorporate spatially localized anomaly synthesis and patch-wise contrastive training.

\begin{table}[!ht]
    \centering
    \scriptsize
    \setlength{\tabcolsep}{0pt}
    \caption{AUROC (\%) comparison across different contrastive objectives for CIFAR-10.}\label{tab:cipher10_per_class_classifiers}
    \begin{tabularx}{\linewidth}{@{} XXXXXXX @{}}
        \toprule
        \textbf{Label} & \textbf{NT-Xent} & \textbf{SupCon} & \textbf{Rot-SupCon} & \textbf{FIRM} & \textbf{FIRM} & \textbf{FIRM} \\
        \midrule
        \textbf{Score} & $\scon $ & $\scon $ & $\scon $ & $\scon $ & $\sshift $ & $\sens $ \\
        \midrule
        0 & 91.7\sd{0.4} & 84.8\sd{0.0} & 86.6\sd{0.0} & 89.2\sd{0.5} & 91.9\sd{0.0} & \textbf{92.6\sd{0.0}} \\
        1 & 98.9\sd{0.0} & 97.8\sd{0.1} & 98.2\sd{0.1} & 98.3\sd{0.0} & 99.1\sd{0.0} & \textbf{99.2\sd{0.0}} \\
        2 & 89.8\sd{0.2} & 65.8\sd{0.5} & 90.3\sd{0.1} & 91.6\sd{0.0} & 93.3\sd{0.1} & \textbf{93.9\sd{0.1}} \\
        3 & 83.4\sd{0.1} & 63.7\sd{0.2} & 80.3\sd{0.4} & 84.0\sd{0.7} & 86.9\sd{0.1} & \textbf{87.6\sd{0.1}} \\
        4 & 91.7\sd{0.2} & 90.5\sd{0.2} & 92.8\sd{0.3} & 93.7\sd{0.0} & 95.0\sd{0.0} & \textbf{95.4\sd{0.0}} \\
        5 & 90.5\sd{0.2} & 88.5\sd{0.3} & 92.2\sd{0.3} & 92.8\sd{0.3} & 94.0\sd{0.0} & \textbf{94.2\sd{0.0}} \\
        6 & 94.6\sd{0.4} & 91.9\sd{0.1} & 94.9\sd{0.0} & 94.8\sd{0.3} & 97.0\sd{0.1} & \textbf{97.4\sd{0.1}} \\
        7 & 98.0\sd{0.0} & 97.1\sd{0.0} & 98.1\sd{0.1} & 98.1\sd{0.0} & 98.8\sd{0.0} & \textbf{98.8\sd{0.0}} \\
        8 & 95.7\sd{0.3} & 95.1\sd{0.1} & 95.9\sd{0.1} & 96.6\sd{0.1} & 97.7\sd{0.0} & \textbf{97.7\sd{0.0}} \\
        9 & 95.6\sd{0.2} & 94.7\sd{0.0} & 95.5\sd{0.0} & 95.3\sd{0.0} & \textbf{96.7\sd{0.0}} & 96.5\sd{0.0} \\
        \midrule
        \textbf{Mean} & 93.0\sd{0.2} & 87.0\sd{0.2} & 92.5\sd{0.2} & 93.4\sd{0.2} & 95.0\sd{0.1} & \textbf{95.3\sd{0.0}} \\ 
\bottomrule
    \end{tabularx}
\end{table}

\begin{table}[!ht]
    \centering
    \scriptsize
    \setlength{\tabcolsep}{0pt}
    \caption{AUROC (\%) comparison across different contrastive objectives for CIFAR-100.}\label{tab:cipher100_per_class_classifiers}
    \begin{tabularx}{\linewidth}{@{}XXXXXXX@{}}
        \toprule
        \textbf{Label} & \textbf{NT-Xent} & \textbf{SupCon} & \textbf{Rot-SupCon} & \textbf{FIRM} & \textbf{FIRM} & \textbf{FIRM} \\
        \midrule
        \textbf{Score} & $\scon $ & $\scon $ & $\scon $ & $\scon $ & $\sshift $ & $\sens $ \\
        \midrule
        0 & 84.8\sd{0.1} & 78.6\sd{0.4} & 81.3\sd{0.2} & 83.9\sd{0.2} & 87.1\sd{0.0} & \textbf{87.2\sd{0.5}} \\
        1 & \textbf{86.9\sd{0.1}} & 75.6\sd{0.0} & 78.9\sd{0.4} & 82.9\sd{0.0} & 86.6\sd{0.3} & 86.7\sd{0.3} \\
        2 & \textbf{93.4\sd{0.3}} & 79.3\sd{1.1} & 85.1\sd{0.9} & 89.0\sd{0.1} & 93.0\sd{0.3} & 93.1\sd{0.4} \\
        3 & 88.3\sd{0.5} & 72.6\sd{10.6} & 88.9\sd{0.2} & 91.1\sd{0.2} & \textbf{92.5\sd{0.4}} & 91.5\sd{0.7} \\
        4 & \textbf{92.8\sd{0.3}} & 81.7\sd{0.1} & 87.5\sd{0.5} & 89.0\sd{0.0} & 92.6\sd{0.2} & 92.7\sd{0.1} \\
        5 & 83.6\sd{0.7} & 57.2\sd{0.8} & 77.6\sd{0.6} & 82.8\sd{0.1} & 86.0\sd{0.2} & \textbf{86.5\sd{0.2}} \\
        6 & 82.2\sd{0.3} & 89.5\sd{0.5} & 90.8\sd{0.0} & 92.4\sd{0.5} & 93.9\sd{0.2} & \textbf{94.3\sd{0.1}} \\
        7 & \textbf{87.2\sd{0.2}} & 70.8\sd{1.1} & 75.5\sd{0.6} & 77.8\sd{0.7} & 84.7\sd{0.0} & 85.4\sd{0.4} \\
        8 & 86.8\sd{0.1} & 87.0\sd{0.1} & 89.1\sd{0.5} & 89.5\sd{0.0} & 91.5\sd{0.2} & \textbf{92.6\sd{0.0}} \\
        9 & 93.4\sd{0.3} & 93.0\sd{0.3} & 93.4\sd{0.2} & 94.0\sd{0.0} & 95.5\sd{0.0} & \textbf{95.9\sd{0.0}} \\
        10 & 87.3\sd{1.0} & 87.9\sd{0.7} & 88.5\sd{0.3} & 88.9\sd{0.1} & 90.7\sd{0.0} & \textbf{90.9\sd{0.0}} \\
        11 & 86.2\sd{0.3} & 86.6\sd{0.2} & 88.7\sd{0.3} & 89.6\sd{0.2} & \textbf{91.3\sd{0.2}} & 91.3\sd{0.0} \\
        12 & 85.3\sd{0.1} & 86.1\sd{0.2} & 87.4\sd{0.0} & 89.0\sd{0.2} & 90.9\sd{0.1} & \textbf{91.4\sd{0.0}} \\
        13 & 81.8\sd{0.9} & 63.8\sd{0.4} & 70.2\sd{1.6} & 76.7\sd{0.6} & 81.7\sd{0.1} & \textbf{82.9\sd{0.0}} \\
        14 & 92.4\sd{0.2} & 92.0\sd{0.1} & 93.2\sd{0.2} & 94.8\sd{0.2} & 96.2\sd{0.3} & \textbf{96.7\sd{0.2}} \\
        15 & 76.8\sd{0.2} & 73.4\sd{0.1} & 76.0\sd{0.2} & 78.2\sd{0.2} & 81.9\sd{0.3} & \textbf{82.6\sd{0.2}} \\
        16 & 81.6\sd{0.3} & 66.6\sd{1.4} & 81.9\sd{0.6} & 83.9\sd{0.3} & 85.9\sd{0.3} & \textbf{86.4\sd{0.4}} \\
        17 & 97.4\sd{0.0} & 95.3\sd{0.2} & 96.1\sd{0.0} & 96.4\sd{0.0} & 98.1\sd{0.1} & \textbf{98.4\sd{0.0}} \\
        18 & 93.6\sd{0.0} & 93.1\sd{0.4} & 94.2\sd{0.1} & 94.8\sd{0.1} & 96.1\sd{0.2} & \textbf{96.4\sd{0.1}} \\
        19 & 93.0\sd{0.0} & 92.3\sd{0.2} & 92.8\sd{0.0} & 93.7\sd{0.1} & 96.0\sd{0.0} & \textbf{96.3\sd{0.1}} \\
        \midrule
        \textbf{Mean} & 87.7\sd{0.3} & 81.1\sd{0.9} & 85.8\sd{0.4} & 87.9\sd{0.2} & 90.6\sd{0.2} & \textbf{91.0\sd{0.2}} \\  
        \bottomrule
    \end{tabularx}
\end{table}

\begin{table}[!ht]
    \centering
    \scriptsize
    \setlength{\tabcolsep}{0pt}
    \caption{AUROC (\%) comparison across different contrastive objectives for Fashion-MNIST.}\label{tab:fmnist_per_class_classifiers}
    \begin{tabularx}{\linewidth}{@{}XXXXXXX@{}}
\toprule
    \textbf{Label} & \textbf{NT-Xent} & \textbf{SupCon} & \textbf{Rot-SupCon} & \textbf{FIRM} & \textbf{FIRM} & \textbf{FIRM} \\
    \midrule
    \textbf{Score} & $\scon $ & $\scon $ & $\scon $ & $\scon $ & $\sshift $ & $\sens $ \\
    \midrule
0 & 94.6\sd{0.1} & \textbf{96.6\sd{0.1}} & 96.4\sd{0.0} & 96.3\sd{0.0} & 96.0\sd{0.1} & 96.0\sd{0.0} \\
1 & 99.4\sd{0.0} & 99.7\sd{0.0} & 99.8\sd{0.0} & 99.8\sd{0.0} & \textbf{99.8\sd{0.0}} & 99.8\sd{0.0} \\
2 & 94.8\sd{0.0} & 95.4\sd{0.2} & 95.5\sd{0.1} & 95.3\sd{0.2} & \textbf{95.5\sd{0.0}} & 95.3\sd{0.1} \\
3 & 94.3\sd{0.2} & 96.5\sd{0.3} & \textbf{96.6\sd{0.4}} & 96.5\sd{0.2} & 95.6\sd{0.0} & 96.0\sd{0.0} \\
4 & 92.7\sd{0.1} & 95.0\sd{0.0} & 94.9\sd{0.0} & \textbf{95.5\sd{0.0}} & 93.7\sd{0.1} & 93.1\sd{0.2} \\
5 & 95.8\sd{0.0} & 97.0\sd{0.2} & 98.1\sd{0.1} & \textbf{98.6\sd{0.1}} & 97.1\sd{0.0} & 96.8\sd{0.0} \\
6 & \textbf{89.0\sd{0.1}} & 86.1\sd{0.3} & 87.5\sd{0.4} & 87.7\sd{0.1} & 88.7\sd{0.2} & 88.4\sd{0.1} \\
7 & 98.8\sd{0.1} & 99.5\sd{0.0} & 99.5\sd{0.0} & \textbf{99.6\sd{0.0}} & 99.2\sd{0.0} & 99.3\sd{0.0} \\
8 & 99.0\sd{0.1} & 99.1\sd{0.0} & 99.5\sd{0.0} & 99.5\sd{0.1} & 99.7\sd{0.0} & \textbf{99.7\sd{0.0}} \\
9 & 99.3\sd{0.0} & 99.4\sd{0.1} & 99.2\sd{0.1} & \textbf{99.5\sd{0.0}} & 99.3\sd{0.0} & 99.3\sd{0.0} \\
 \midrule
\textbf{Mean} & 95.8\sd{0.1} & 96.4\sd{0.1} & 96.7\sd{0.1} & \textbf{96.8\sd{0.1}} & 96.5\sd{0.1} & 96.4\sd{0.1} \\
    \bottomrule
    \end{tabularx}
\end{table}

\begin{table}[!ht]
    \centering
    \scriptsize
    \setlength{\tabcolsep}{0pt}
    \caption{AUROC (\%) comparison across different contrastive objectives for Cats-vs-Dogs.}\label{tab:catsdogs_per_class_classifiers}
    \begin{tabularx}{\linewidth}{@{}XXXXXXX@{}}
    \toprule
    \textbf{Label} & \textbf{NT-Xent} & \textbf{SupCon} & \textbf{Rot-SupCon} & \textbf{FIRM} & \textbf{FIRM} & \textbf{FIRM} \\
    \midrule
    \textbf{Score} & $\scon $ & $\scon $ & $\scon $ & $\scon $ & $\sshift $ & $\sens $ \\
    \midrule
    0 & \textbf{91.2\sd{0.0}} & 57.3\sd{0.7} & 87.9\sd{0.2} & 89.8\sd{0.7} & 89.6\sd{0.3} & 89.8\sd{0.3} \\
    1 & 85.0\sd{0.3} & 59.1\sd{0.3} & 90.4\sd{0.1} & \textbf{91.1\sd{0.2}} & 90.3\sd{0.3} & 90.5\sd{0.2} \\
    \midrule
    \textbf{Mean} & 88.1\sd{0.2} & 58.2\sd{0.5} & 89.2\sd{0.1} & \textbf{90.4\sd{0.5}} & 89.9\sd{0.3} & 90.2\sd{0.3} \\ 
    \bottomrule
    \end{tabularx}
\end{table}

\begin{table*}[!t]
    \centering
    \scriptsize
    \setlength{\tabcolsep}{4pt}
    \caption{{Further comparison AUROC (\%) for image-level anomaly detection for MVTec-AD. These results are from the ablation study where FIRM is trained with NSA and whole images, without the SIA and patch-based strategy.}}
    \label{tab:results_mvtec_perclass_full}
    \begin{tabularx}{\linewidth}{@{}lXXXXXXXX@{}}
        \toprule
        \textbf{Method} & \textbf{RotNet} & \textbf{DROC} & \textbf{DOCC} & \textbf{CutPaste} & \textbf{P-SVDD} & \textbf{U-Student} & \textbf{NSA} & \textbf{FIRM} \\
        \midrule
        Bottle       & --    & --    & 99.6 & 99.2\sd{0.2} & 98.6 & 96.7 & 97.6\sd{0.2} & \textbf{100\sd{0.0}} \\
        Cable        & --    & --    & 90.9 & 87.1\sd{0.8} & 90.3 & 82.3 & 92.1\sd{2.4} & \textbf{97.4\sd{0.3}} \\
        Capsule      & --    & --    & 91.0 & 87.9\sd{0.7} & 76.7 & 92.8 & \textbf{93.2\sd{0.8}} & 92.8\sd{0.2} \\
        Hazelnut     & --    & --    & 95.0 & 91.3\sd{0.6} & 92.0 & 91.4 & 93.5\sd{1.9} & \textbf{96.5\sd{0.1}} \\
        Metal Nut    & --    & --    & 85.2 & 96.8\sd{0.5} & 94.0 & 94.0 & \textbf{99.4\sd{0.3}} & 98.6\sd{0.2} \\
        Pill         & --    & --    & 80.4 & 93.4\sd{0.9} & 86.1 & 86.7 & \textbf{97.0\sd{0.9}} & 92.8\sd{0.6} \\
        Screw        & --    & --    & 86.9 & 93.4\sd{0.9} & 81.3 & 87.4 & 90.3\sd{1.2} & \textbf{96.7\sd{0.5}} \\
        Toothbrush   & --    & --    & 96.4 & 99.2\sd{0.2} & \textbf{100} & 98.6 & \textbf{100\sd{0.0}} & \textbf{100\sd{0.0}} \\
        Transistor   & --    & --    & 90.8 & \textbf{96.4\sd{0.7}} & 91.5 & 83.6 & 93.5\sd{0.9} & 93.1\sd{0.3} \\
        Zipper       & --    & --    & 92.4 & 99.4\sd{0.1} & 97.9 & 95.8 & 99.8\sd{0.1} & \textbf{100\sd{0.0}} \\
        \midrule
        \, \textbf{Objects Mean} &  -- & -- & 90.8 & 94.4\sd{0.5} & 90.8 & 90.9 & 95.6\sd{0.9} & \textbf{96.8\sd{0.2}} \\ 
        \midrule
        Carpet       & --    & --    & 90.6 & 67.9\sd{1.8} & 92.9 & \textbf{95.3 }& 85.6\sd{7.6} & 75.2\sd{0.3} \\
        Grid         & --    & --    & 52.4 & 99.9\sd{0.1} & 94.6 & 98.7 & 99.9\sd{0.1} & \textbf{100\sd{0.0}} \\
        Leather      & --    & --    & 78.3 & 99.7\sd{0.1} & 90.9 & 93.4 & \textbf{99.9\sd{0.1}} & 93.8\sd{0.8} \\
        Tile         & --    & --    & 96.5 & 95.9\sd{1.0} & 97.8 & 95.8 & 99.7\sd{0.2} & \textbf{100\sd{0.0}} \\
        Wood         & --    & --    & 91.6 & 94.9\sd{0.5} & 96.5 & 95.5 & \textbf{96.7\sd{1.2}} & 87.5\sd{0.3} \\
        \midrule
        \,\textbf{Texture Mean} &  -- & -- & 81.9 & 91.7\sd{0.7} & 94.5 & 95.7 & 96.3\sd{1.8} & 91.3\sd{0.28} \\
        \midrule
        \textbf{Overall Mean} &  71.0\sd{3.5} & 86.5\sd{1.6} & 87.9 & 90.9\sd{0.7} & 92.1 & 92.5 & \textbf{95.9\sd{0.7}} & 95.0\sd{0.2} \\
        \bottomrule
    \end{tabularx}
\end{table*}

\end{document}